\documentclass[sigconf]{acmart}
\copyrightyear{2025}
\acmYear{2025}
\setcopyright{acmlicensed}
\acmConference[WWW '25] {Proceedings of the ACM Web Conference 2025}{April 28-May 2, 2025}{Sydney, NSW, Australia.}
\acmBooktitle{Proceedings of the ACM Web Conference 2025 (WWW '25), April 28-May 2, 2025, Sydney, NSW, Australia}
\acmDOI{10.1145/3696410.3714816}
\acmISBN{979-8-4007-1274-6/25/04}

\usepackage{hyperref}
\usepackage{hyperxmp}
\usepackage{amsmath,amsfonts}
\usepackage{multirow}
\usepackage{tabularx}
\usepackage{csquotes}
\usepackage{arydshln}
\usepackage{pifont}%
\usepackage{titlesec}
\usepackage{bbding}
\usepackage{graphicx}

\usepackage{CJKutf8}

\usepackage{balance}
\definecolor{deepgreen}{rgb}{0.0, 0.4, 0.0}
\definecolor{lightblue}{rgb}{0.9, 0.96, 1}
\definecolor{deepred}{rgb}{0.7, 0.0, 0.0}
\newcommand{\model}{OntoTune}

\AtBeginDocument{%
  }

\begin{document}



\title{\model: Ontology-Driven Self-training for Aligning Large Language Models}


\author{Zhiqiang Liu}
\affiliation{%
  \institution{Zhejiang University}
  \city{Hangzhou}
  \country{China}}
\email{zhiqiangliu@zju.edu.cn}

\author{Chengtao Gan}
\affiliation{%
  \institution{Zhejiang University}
  \city{Hangzhou}
  \country{China}}
\email{22451144@zju.edu.cn}

\author{Junjie Wang}
\affiliation{%
  \institution{Zhejiang University}
  \city{Hangzhou}
  \country{China}}
\email{wangjj2018@zju.edu.cn}

\author{Yichi Zhang}
\affiliation{%
  \institution{Zhejiang University}
  \city{Hangzhou}
  \country{China}}
\email{zhangyichi2022@zju.edu.cn}

\author{Zhongpu Bo}
\affiliation{%
  \institution{Ant Group}
  \city{Hangzhou}
  \country{China}}
\email{bozhongpu.bzp@antgroup.com }

\author{Mengshu Sun}
\affiliation{%
  \institution{Ant Group}
  \city{Hangzhou}
  \country{China}}
\email{mengshu.sms@antgroup.com}

\author{Huajun Chen}
\affiliation{%
  \institution{Zhejiang University}
  \city{Hangzhou}
  \country{China}}
\email{huajunsir@zju.edu.cn}

\author{Wen Zhang}
\authornote{Corresponding author}
\affiliation{%
  \institution{Zhejiang University}
  \city{Hangzhou}
  \country{China}}
\email{zhang.wen@zju.edu.cn}







\renewcommand{\shortauthors}{Zhiqiang Liu et al.}

\begin{abstract}
Existing domain-specific Large Language Models (LLMs) are typically developed by fine-tuning general-purposed LLMs with large-scale domain-specific corpora. However, training on large-scale corpora often fails to effectively organize domain knowledge of LLMs, leading to fragmented understanding. Inspired by how humans connect concepts and organize knowledge through mind maps, we aim to emulate this approach by using \textbf{ontology} with hierarchical conceptual knowledge to reorganize LLM's domain knowledge. From this perspective, we propose an ontology-driven self-training framework called \textbf{OntoTune}, which aims to align LLMs with ontology through in-context learning, enabling the generation of responses guided by the ontology. We leverage in-context learning to identify whether the LLM has acquired the specific concept's ontology knowledge, and select the entries not yet mastered by LLM as the training set to further align the LLM with ontology. Compared to existing domain LLMs based on newly collected large-scale domain-specific corpora, our OntoTune, which relies on the existing, long-term developed ontology and LLM itself, significantly reduces data maintenance costs and offers improved generalization ability. We conduct our study in the medical domain to evaluate the effectiveness of OntoTune, utilizing a standardized medical ontology, SNOMED CT as our ontology source. Experimental results demonstrate that OntoTune achieves state-of-the-art performance in both in-ontology task hypernym discovery and out-of-ontology task medical domain QA. Moreover, compared to the latest direct ontology injection method TaxoLLaMA, our OntoTune better preserves original knowledge of LLM. The code and data are available at \url{https://github.com/zjukg/OntoTune}.
\end{abstract}






\maketitle

\section{Introduction}
\begin{CJK}{UTF8}{gbsn}
Large Language Models (LLMs), such as GPT-4 \cite{DBLP:journals/corr/abs-2303-08774} and LLaMA \cite{dubey2024llama}, have achieved remarkable success in the field of natural language processing \cite{DBLP:journals/corr/abs-2407-15017}, demonstrating advanced performance across various domains and tasks. 
To further enhance the capabilities of LLMs in specific domain, such as medical, financial, and science, the research and industry community have begun to focus on developing domain-specific LLMs \cite{DBLP:conf/acl/LabrakBMGRD24,DBLP:conf/acl/BhatiaNCA24,DBLP:journals/ai/AlmeidaNEWA24}. 

Existing methods usually develop domain-specific LLMs by further training general-purposed LLMs on domain-specific corpora, such as BloombergGPT \cite{DBLP:journals/corr/abs-2303-17564}, BioMistral \cite{DBLP:conf/acl/LabrakBMGRD24} and LawGPT \cite{DBLP:journals/corr/abs-2406-04614}. Previous researches \cite{DBLP:conf/acl/RenCLLHZWC024,DBLP:journals/corr/abs-2405-05904} indicate that LLMs have already acquired most domain knowledge during the comprehensive pre-training phase, and need to reorganize and align knowledge with domain-specific requirements during the post-training phase. However, adapting LLMs to specific domains presents significant challenges \cite{DBLP:journals/corr/abs-2311-05112,DBLP:conf/acl/LiYBZLSLSYWLXBF24}. On the one hand, the scarcity of domain-specific corpora and constraints imposed by data privacy present significant hurdles in the continuous collection of high-quality domain corpora for continual pre-training or supervised fine-tuning, demanding substantial investment in time and resources. On the other hand, existing researches \cite{DBLP:conf/iclr/ChengHW24,dorfner2024biomedical} reveal that directly fine-tuning LLMs with fragmented raw domain corpora struggles to effectively organize domain knowledge and can even impair prompting capabilities of LLMs. \textbf{So can we find a more efficient alternative to reorganize domain knowledge in large language models without relying on large-scale domain-specific corpora?}

\begin{figure}
\centering
\includegraphics[scale=0.56]{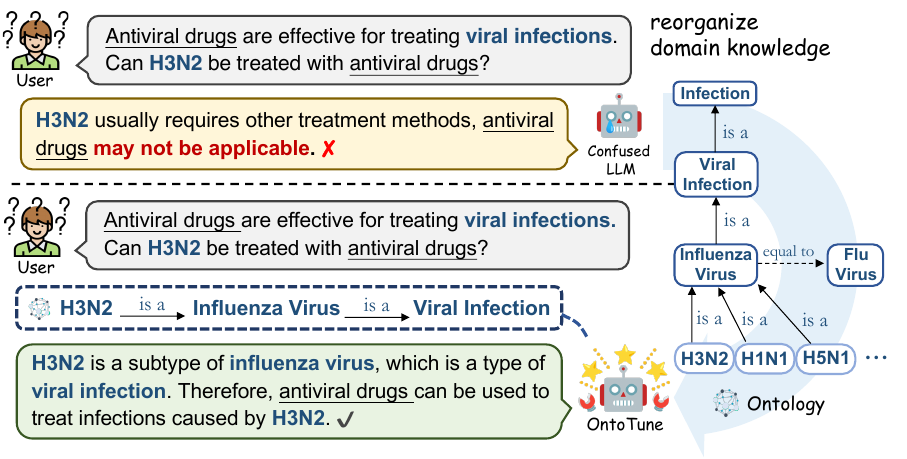}
\vspace{-7mm}
\caption{A simple example illustrates how hierarchical structure knowledge in the ontology guide responses.}
\vspace{-3mm}
\label{fig:intro}
\end{figure}

\textit{Inspired by how humans use mind maps 
which visually represent concepts and their relationships, to systematically organize and review knowledge, we aim to use domain-specific mind maps to reorganize LLM's domain knowledge.} Naturally, we associate these mind maps with widely established, rigorously constructed \textbf{ontologies} \cite{DBLP:conf/ijcai/XiaoCKLPRZ18}, which fully display the relationships and hierarchical structures between domain concepts as the ideal domain-specific mind maps. As shown in Figure \ref{fig:intro}, the ontology structure primarily consists of hypernym and synonym relationships between concepts, and have been widely applied in scenarios such as information retrieval \cite{DBLP:journals/corr/abs-2407-10671} and knowledge reasoning \cite{DBLP:conf/www/ZhangPWCZZBC19,DBLP:conf/acl/0002WHZS0WLFS023}. Common domain ontologies include SNOMED CT \cite{DBLP:journals/midm/SchulzK08} in the biomedical field, WordNet \cite{DBLP:conf/naacl/Miller94a} in the lexical field and GeoNames\footnote{https://www.geonames.org/} in the geographical field. Figure \ref{fig:intro} illustrates an example of a medical ontology guided response, where the LLM links concepts through the hierarchical structure knowledge in the ontology. Meanwhile, we suppose that compared to collecting new large-scale domain corpora, utilizing existing, long-term developed ontologies can reduce data maintenance costs and offer better generalization. From this perspective, we propose an \underline{onto}logy-driven self-training fine-\underline{tun}ing fram\underline{e}work \textbf{OntoTune}, which aims to align LLMs with domain ontology through in-context learning\footnote{https://openai.com/index/learning-to-reason-with-llms/} and generate responses guided by the ontology. OntoTune's workflow consists of three main steps: \textbf{(1) Instruction Text Generation.} We utilize three ontology-aware concept-level instructions which focus respectively on diversity, conceptuality, and professionalism to generate outputs. Then we incorporate the corresponding ontology knowledge to the input and let seed model rethink to obtain better outputs through in-context learning. \textbf{(2) Inconsistency Text Selection.} If there is significant inconsistency between the corpora obtained with and without ontology knowledge, we consider that the seed model has not effectively grasped this concept's ontology structure to guide its output and select entries that exhibit significant inconsistency as the training set. \textbf{(3) LLM Fine-tuning.} Based on the training set, we perform self-training on the seed model, resulting in aligned domain LLMs.

We conduct our study in the medical field, using the high-quality medical ontology SNOMED CT \cite{DBLP:journals/midm/SchulzK08} as the source ontology. To evaluate the effectiveness of  \model, we compare it not only with customized models for specific tasks but also with existing domain LLM trained on large-scale corpora and the direct ontology injection method TaxoLLaMA* \cite{DBLP:conf/acl/MoskvoretskiiNL24} implemented on the same LLM called seed model. Results show that we have achieved state-of-the-art performance in in-ontology task hypernym discovery and out-of-ontology task domain QA, demonstrating that OntoTune can effectively improve the performance of domain-specific tasks. Moreover, OntoTune significantly preserves the knowledge and safety of the seed model compared to existing domain-specific LLMs and TaxoLLaMA. Our contributions can be summarized as follows:
\begin{list}{\huge\textbullet}{\leftmargin=1.5em}
\item We highlight the limitations of developing domain LLMs based on large-scale domain corpora, and we are the first to utilize small-scale ontology to reorganize the domain knowledge of LLMs.
\item We propose a novel ontology-driven self-training method OntoTune, which aligns LLMs with ontologies through in-context learning, thereby guiding LLMs to generate responses under domain ontology knowledge.

\item Compared to exsiting domain LLM based on large-scale raw domain corpora and the direct injection method TaxoLLaMA, our OntoTune achieves state-of-the-art performance in the in-ontology task hypernym discovery and out-of-ontology task domain QA, and significantly preserves the knowledge capabilities and safety of the seed model.
\end{list}
\section{Related Works}

\begin{figure*}[ht]
\centering
\includegraphics[scale=0.67]{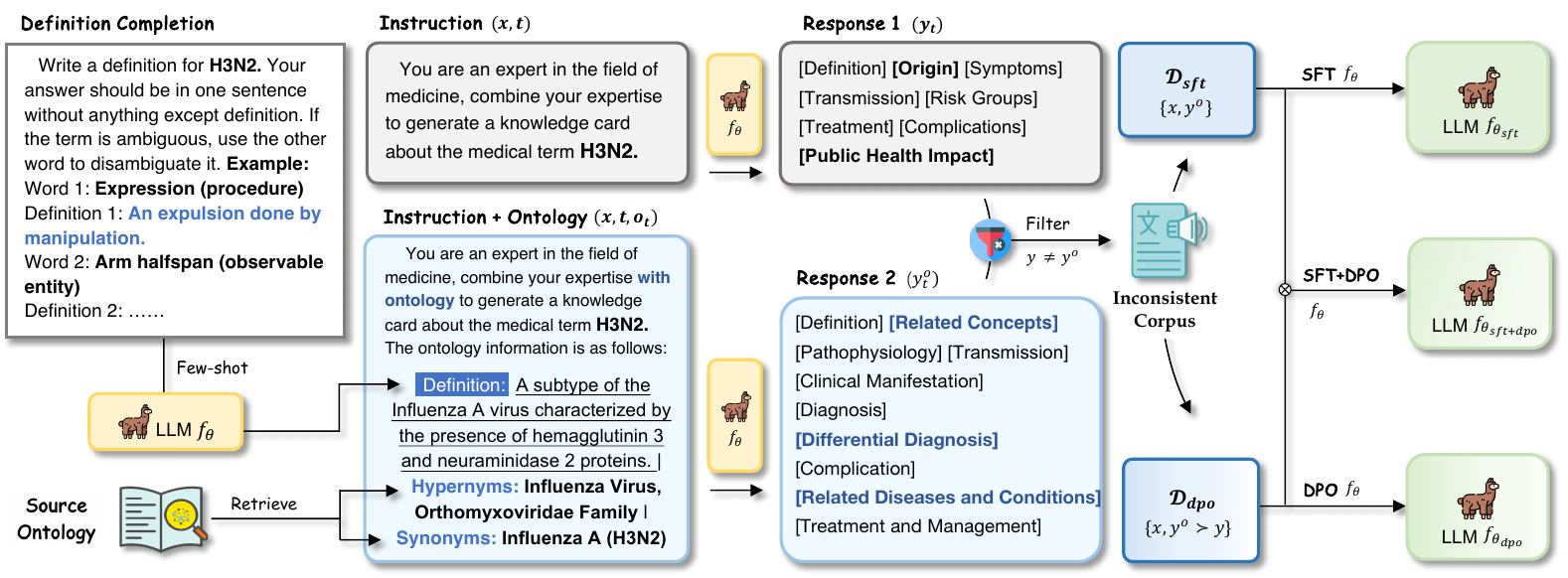}
\vspace{-6mm}
\caption{Overview of OntoTune which aligns LLMs with ontology through in-context learning.}
\label{fig:method}
\end{figure*}

\subsection{Domain-specific LLMs}
Existing domain-specific large language models (LLMs) can be categorized into two groups: (1) those models trained from scratch using domain-specific corpora, such as BioGPT \cite{DBLP:journals/bib/LuoSXQZPL22} and GatorTron \cite{DBLP:journals/corr/abs-2203-03540}, and (2) those \cite{DBLP:conf/icde/00070HCGYBZYSWY23,DBLP:conf/acl/ZhangCFLL0C24,DBLP:conf/acl/LabrakBMGRD24, DBLP:journals/corr/abs-2303-17564} that employ continual training on general-purposed models. Benefiting from its ability to leverage the extensive and diverse data of the seed models, as well as more efficient training processes, the latter approach has gradually become mainstream. Current domain-specific LLMs like BioMistral \cite{DBLP:conf/acl/LabrakBMGRD24}, BloombergGPT \cite{DBLP:journals/corr/abs-2303-17564} and LawGPT \cite{DBLP:journals/corr/abs-2406-04614} are developed by training a seed model with a large-scale raw domain-specific corpora, demonstrating impressive performance on domain tasks. To be specific, the medical model PMC-LLaMA \cite{DBLP:journals/corr/abs-2304-14454} is fine-tuned with LoRA \cite{DBLP:conf/iclr/HuSWALWWC22}  on LLaMA using 4.8 million biomedical papers. LawGPT \cite{DBLP:journals/corr/abs-2406-04614}  continues training on 500k legal documents. And BloombergGPT \cite{DBLP:journals/corr/abs-2303-17564} is fine-tuned on a 708 billion tokens financial corpora. These models typically rely on large amounts of training data to adapt to their respective domains. However, this fragmented knowledge from the raw corpora is merely injected into the seed model without being systematically organized and recent research \cite{DBLP:conf/iclr/ChengHW24,dorfner2024biomedical} have indicated that directly using these fragmented raw corpora is not efficient. Additionally, prior researches seldom utilize ontologies as foundational knowledge sources for training corpora. Compared to fragmented large-scale corpora, concept-level structured knowledge in ontologies can play a significant role in knowledge management \cite{DBLP:journals/corr/abs-2407-10671} and semantic search \cite{DBLP:conf/www/ZhangPWCZZBC19,DBLP:conf/acl/0002WHZS0WLFS023}, and also have the potential to empower LLMs. Recently, TaxoLLaMA \cite{DBLP:conf/acl/MoskvoretskiiNL24} develops a lexical semantic LLM via directly employing the WordNet \cite{DBLP:conf/naacl/Miller94a} ontology for instruction-tuning, achieving state-of-the-art performance in multiple lexical semantic tasks and highlighting the potential of ontologies for developing domain-specific LLMs.

\subsection{Self-Generated Data for Training}
The self-training paradigm involves generating data autonomously and using this self-generated data for further training. Traditional self-training methods \cite{DBLP:conf/iclr/HeGSR20,DBLP:conf/cvpr/XieLHL20,DBLP:journals/corr/abs-2202-12040,DBLP:conf/emnlp/0001GHW00023} typically employ a trained model to annotate data, and then improve model performance based on these newly annotated data. Due to its simplicity and efficiency, this training paradigm is also migrating to LLMs. Given the high costs of manually annotating training data or using more powerful proprietary models like GPT-4 \cite{DBLP:journals/corr/abs-2303-08774}, many works \cite{DBLP:conf/icml/MengMHZA023,DBLP:conf/acl/YangPFWCZL24,DBLP:conf/acl/WangKMLSKH23,DBLP:journals/corr/abs-2402-06457,DBLP:journals/tmlr/SinghCAAPGLH0XP24} have begun to leverage the language model itself to synthesize training data. STaR \cite{DBLP:conf/nips/ZelikmanWMG22} is a self-taught reasoner that learns from its own generated reasoning steps to improve reasoning ability. Furthermore, SDFT \cite{DBLP:conf/acl/YangPFWCZL24} proposes a self-distillation fine-tuning method to achieve more efficient and less damaging results. Alternatively, Lin et al. \cite{DBLP:journals/corr/abs-2404-07965} use gold answers to train a reward model for evaluating generated instructions separately. However, previous self-training approaches usually rely on gold labels to filter out low-quality instruction data, and they tend to focus more on improvements within a single dataset. Unlike previous methods, our OntoTune mitigates performance degradation caused by incorrect labels by refining and reorganizing internal domain knowledge of the seed model through open-ended instructions \cite{DBLP:conf/iclr/0009CMZYSZ24,DBLP:conf/acl/TyenMCCM24}.

\section{Methodology}
In this section, we first set an objective to evaluate whether the seed model has mastered domain ontology knowledge and guide the model's responses. To achieve this objective, we introduce an \textbf{Onto}logy-driven self-training fine-\textbf{tun}ing fram\textbf{e}work \textbf{OntoTune}.

\subsection{Objective Defintion}

Given an instruction $x$  that is closely related to ontology knowledge $o$, we could get two kinds of responses:
\begin{align}
    y = f(x) \quad\text{and}\quad y^o  = f(x, o),
\end{align}
where $y$ is the response with instruction $x$ as input, and the $y^o$ is the response with both instruction $x$ and the ontology knowledge $o$ as input. 
We hypothesize that if the seed model $f$ has fully mastered and properly utilizes the ontology knowledge when generating response, then $y$ should equal to $y^o$. Otherwise, $y^o$ should be better than $y$, since LLMs have the in-context-learning capability, and the inclusion of $o$ could lead to more systematic and logical responses. However, from our experience, $y$ is not close to or similar to $y^o$ in a lot of cases, which can be found in Appendix \ref{app:d}. 

To internalize the ontology knowledge into to LLMs, we align seed model $f_{\theta}$, which has parameter $\theta$ , to ontology through instruction tuning, getting model $f_{\theta'}$ with updated parmeters $\theta'$. 
We establish the optimization objective to 
\begin{equation}
    f_{\theta'}(x) = f_{\theta}(x, o)
\end{equation}
As analyzed before, this objective approximately means $f_{\theta'}$ has mastered the ontology knowledge and could properly utilize the internal ontology knowledge when generating response. 

\subsection{\model}

To effectively internalize ontology knowledge,, we introduce the \textbf{OntoTune} framework as shown in Fig \ref{fig:method}. 
The OntoTune workflow consists of three main steps: \textbf{(1) Instruction text generation.} We utilize three types of concept-level ontology-aware instructions that include (or exclude) ontology knowledge as input to the seed model. These instructions focus on diversity, conceptuality, and professionalism. \textbf{(2) Inconsistency text selection.} We select responses that exhibit significant inconsistency between those that include and those that exclude ontology knowledge as our training set. \textbf{(3) LLM Fine-tuning.} Based on training set, we perform self-training on the seed model.

Previous researches point \cite{DBLP:conf/acl/MoskvoretskiiNL24,DBLP:conf/coling/MoskvoretskiiPN24} that the definitions of concepts are crucial for ontology learning tasks. Considering that our framework aims to employ a self-training approach, rather than distilling knowledge from more advanced models like GPT-4 \cite{DBLP:journals/corr/abs-2303-08774}. Therefore, we use the seed model to complete the missing definitions in the ontology via a few-shot learning approach. We provide relevant domain concepts with their definitions as examples and the specific prompt template is shown in Figure \ref{fig:method}.

\subsubsection{\textbf{Instruction Text Generation.}}
\begin{figure}
\centering
\includegraphics[scale=0.541]{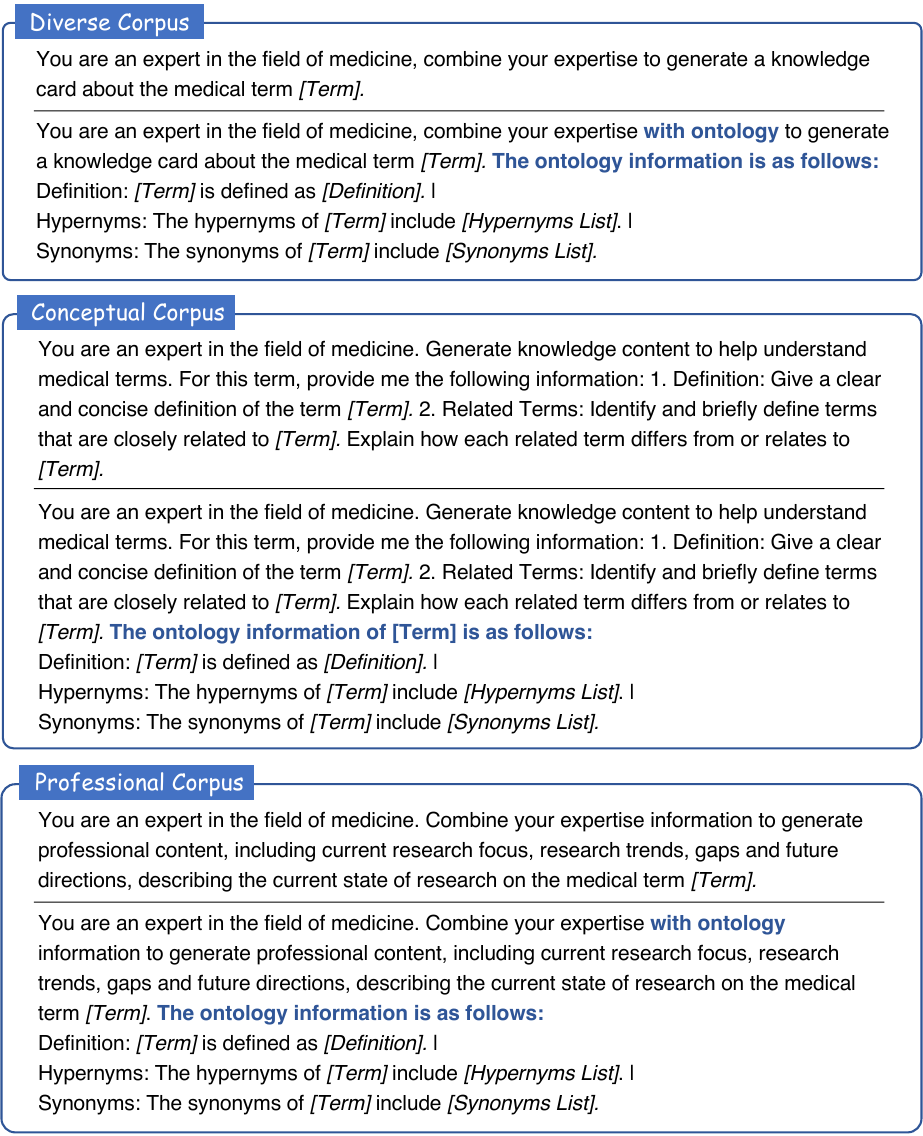}
\vspace{-6mm}
\caption{Ontology-aware corpus generation templates.}
\vspace{-2mm}
\label{fig:prompt}
\end{figure}

To assess to what extent LLMs comprehend ontology knowledge across various dimensions, we design three distinct concept-level instruction templates. These templates evaluate whether the ontology knowledge embedded in the seed model can effectively guide the responses from the perspectives of diversity, conceptuality and professionalism:
\begin{list}{\huge\textbullet}{\leftmargin=1.5em}
\item\textbf{Diverse corpus $x_d$.} This template requires to generate knowledge cards related to specific concepts. The concept's knowledge card is a concise collection of information about a specific domain concept, typically including its definition, related concepts, usage examples, and other supplementary information. 
\item\textbf{Conceptual corpus $x_c$.} This template is directly related to ontology concepts. It requires to generate definitions for concepts and distinguish between related concepts. Ontology can directly guide the model in systematically organizing and describing various concepts and their relationships. 
\item\textbf{Professional corpus $x_p$.} This template requires to elucidate the current research status of the concept in existing academic journals. Ontology implicitly connects related concepts, allowing for a more comprehensive and coherent presentation of academic knowledge.
\end{list}
These corpus generation templates are shown in Figure \ref{fig:prompt}. For the concept $t$, we denote the concept-level instructions as $x\in\{x_d,x_c,x_p\}$, and the generation process is represented as:
\begin{equation}
    y_t=f_\theta(x,t)
\end{equation}
Aiming to align seed model with ontology through in-context learning, we integrate ontology information related to concepts into the input and obtain more systematic and semantically clear responses under the guidance of ontology as shown in Figure\ref{fig:method}. The ontology information includes the \underline{definitions} of concepts and the ontology structure of the concepts, i.e., their \underline{hypernyms} and \underline{synonyms}, which are retrieved from the source ontology. We represent the generation process with concept's ontology information as:
\begin{equation}
    y^o_t=f_\theta(x,t,o_t)
\end{equation}
where $o_t\in O$ is the ontology information about the concept $t$ retrieved from the source ontology $O$ or completed by seed model.

\subsubsection{\textbf{Inconsistency Text Selection.}}
For the concept $t$, if the responses $y_t$ and $y_t^o$ are consistent, it indicates that ontology knowledge related to concept $t$ embedded in the seed model can implicitly guide the response. Conversely, if there is an inconsistency as shown in the example in Figure \ref{fig:method}, the content in $y_t$ is broad but superficial and does not involve related concepts, whereas the content in $y^o_t$ is specific and connected to relevant ontology concepts. Therefore, we select the inconsistent responses as the training set for the seed model to align with ontology. To evaluate inconsistency, we calculate a hybrid similarity score based on three different metrics: embedding cosine similarity, ROUGE-L, and BLEU-4:
\begin{equation}
\begin{aligned}
    {\rm sim}(y_t,y^o_t)=&\frac{E^{\top}(y_t)E(y^o_t)}{\|E(y_t)\|\|E(y^o_t)\|}\\&+\text{ROUGE-L}(y_t,y^o_t)+\text{BLEU-4}(y_t,y^o_t)   
\end{aligned}
\end{equation}
where $E(\cdot)$ is a sentence encoding model that encodes the input sentence into a vector for semantic similarity evaluation, which is a fine-tuned model based on MiniLMv2 \cite{DBLP:conf/acl/WangBHDW21} implemented by sentence-transformers\footnote{https://github.com/UKPLab/sentence-transformers} during experiments. And $\text{ROUGE-L}(\cdot)$ and $\text{BLEU-4}(\cdot)$ compute word-level text similarity. We select the lowest $k$ entries based on ${\rm sim}(y_t,y^o_t)$ from each type of corpora to construct the training data. Specifically, we construct our train set under two injection methods: supervised fine-tuning (SFT) data denoted as $\mathcal{D}_{sft}=\{x_n,y_n^o\}_{n=1}^k$ and direct preference optimization (DPO) data denoted as $\mathcal{D}_{dpo}=\{x_n,y_n^o \succ y_n\}_{n=1}^k$.

\subsubsection{\textbf{LLM Fine-tuning.}}
Based on the training set constructed above, we use three fine-tuning methods: supervised instruction fine-tuning (SFT), direct preference optimization (DPO), and supervised instruction fine-tuning combined with direct preference optimization (SFT+DPO). Through SFT, we hope the seed model can directly learn ontology-guided responses from $y_t^o$, thereby implicitly enhancing its internal ontology knowledge. We utilize the training data $\mathcal{D}_{sft}$ to fine-tune the LLM $f_\theta$ directly with the next-token prediction objective for response $y^o_t$:
\begin{equation}
    \mathop{\max}_{\theta} \mathbb{E}_{\left(x_t,y^o_t\right) \sim \mathcal{D}_{sft}}\left[\log P_{\theta}(y^o_t \mid x_t)\right]
\end{equation}
For DPO, we use this fine-tune approach enables the seed model to favor the responses guided by ontology, avoiding the original superficial ones. We utilize the training data $\mathcal{D}_{dpo}$ to optimize the LLM $f_\theta$ by treating $y^o_t$ as the preferred response and $y_t$ as the rejected response:
\begin{equation}\resizebox{0.92\hsize}{!}{$
    \mathop{\max}_{\theta} \mathbb{E}_{\left(x_t,y^{o}_t\succ y_t\right) \sim \mathcal{D}_{dpo}}\left[\log \sigma\left(\beta \log \frac{P_{\theta}\left(y^{o}_t \mid x_t\right)}{P_{\mathrm{ref}}\left(y^{o}_t \mid x_t\right)}-\beta \log \frac{P_{\theta}\left(y_t \mid x_t\right)}{P_{\mathrm{ref}}\left(y_t \mid x_t\right)}\right)\right]$}
\end{equation}
where ${ref}$ is the parameter of initial seed model and $\beta$ is a parameter controlling the deviation from reference policy $P_{ref}$. 
Lastly, following the paradigm of combining SFT and DPO to enhance the model's task adaptability and domain generalization capabilities in previous work \cite{dubey2024llama,DBLP:journals/corr/abs-2303-08774}, we also attempt to train our seed model in two stages using SFT and DPO fine-tuning methods, respectively.

\begin{figure}[t]
\centering
\includegraphics[scale=0.587]{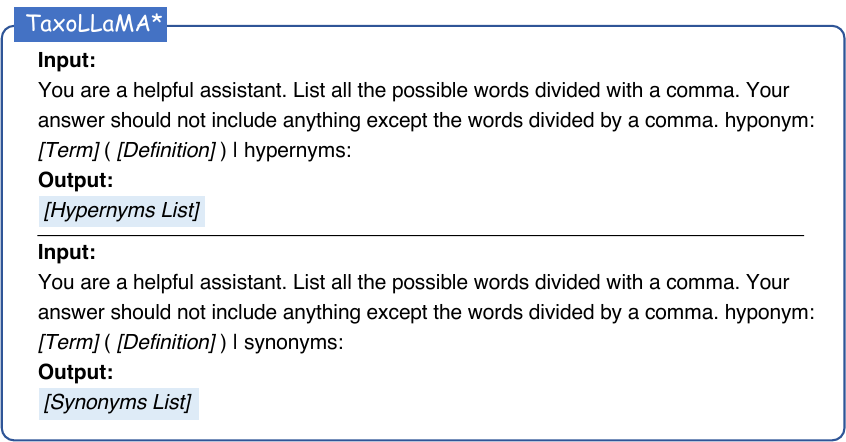}
\vspace{-6mm}
\caption{The templates of TaxoLLaMA*'s instruction-tuning and hypernym discovery task.}
\vspace{-2mm}
\label{fig:taxollama}
\end{figure}

\section{Experiment}

We conduct comprehensive experiments to demonstrate the effectiveness of OntoTune. These experiments are designed to answer the following research questions:
\begin{list}{\huge\textbullet}{\leftmargin=1.5em}
    \item \textbf{RQ1:} Can OntoTune’s implicit injection approach enable LLMs to effectively align with ontology knowledge?
    \item \textbf{RQ2:} Can OntoTune adapt LLMs to specific domains, improving the performance of domain-specific tasks?
    \item \textbf{RQ3:} How does OntoTune affect on the general performance of the seed model?

\end{list}

\subsection{Experimental Setup}
In this paper, we select the medical domain as example to evaluate the effectiveness of our method, since medical field receives widespread attention and has rich evaluation datasets and baselines \cite{DBLP:journals/corr/abs-2311-05112}.  
Specifically, we adopt standardized SNOMED CT\footnote{https://www.snomed.org/} \cite{DBLP:journals/midm/SchulzK08} International Edition June version as our source ontology, which includes 367,978 med
ical concepts, of which only 8,275 have corresponding definitions, and 246,356 taxonomic relationships (i.e., `is-a'). In order to match the training scale of existing domain-specific LLMs \cite{DBLP:journals/corr/abs-2405-01886,DBLP:journals/corr/abs-2408-06142}, we select $k=100000$ inconsistent samples on each type of corpora for training.

We utilize the LLaMA-3-8B-Instruct \cite{dubey2024llama} model as our seed model due to its robustness and generalization across multiple medical tasks. We employ the Low Rank Adaptation \cite{DBLP:conf/iclr/HuSWALWWC22} (LoRA) technique to fine-tune the model based on the LLaMA-Factory \cite{zheng2024llamafactory} framework. During the OntoTune training phase, we apply LoRA to all linear layers with a rank of $r = 8$. All training is conducted on 8 NVIDIA H100 80G GPUs. For SFT stage, we use fp32 and a learning rate of 5e-5, training for 3 epochs with a cosine scheduler, a batch size per device initialized to 8 and gradient accumulation of 2. For DPO stage, we use fp32 and a learning rate of 5e-6, training for 3 epochs with a cosine scheduler and 4 batch size per device.

\begin{table}
\caption{Results of the hypernym discovery. * represent language models that have been adapted for hypernym discovery task. All scores are magnified by a factor of 100.}
\vspace{-2mm}
\centering
\resizebox{0.465\textwidth}{!}{
\begin{tabular}{lcccc}
\toprule
\textbf{Model}    & \multicolumn{1}{l}{\textbf{1A:English}} & \multicolumn{1}{l}{\textbf{1B:Italian}} & \multicolumn{1}{l}{\textbf{1C:Spanish}} & \multicolumn{1}{l}{\textbf{2A:Medical}} \\ \midrule
CRIM \cite{DBLP:conf/semeval/Bernier-Colborne18}             & 36.10                                   & -                                       & -                                       & 54.64                                   \\
Hybird \cite{DBLP:conf/acl/HeldH19}           & 34.07                                   & -                                       & -                                       & \underline{64.47}                                   \\
RMM  \cite{DBLP:conf/acl/BaiZKCM21}              & 39.07                                   & -                                       & -                                       & 54.89                                   \\
300-sparsans \cite{DBLP:conf/semeval/BerendMF18}     & -                                       & 25.14                                   & 37.56                                   & -                                       \\ 
T5$^*$ \cite{DBLP:conf/ijcnlp/NikishinaCDPB23}               & 45.22                                   & 24.04                                   & 27.50                                   & 44.73                                   \\ \midrule
LLaMA3 8B* \cite{dubey2024llama}     & \underline{51.64}                                      & 47.41                                   & 53.06                                  & 54.86                                       \\
Aloe* \cite{DBLP:journals/corr/abs-2405-01886} &45.26	&43.52	&51.03	&57.42 \\ 
Med42-v2* \cite{DBLP:journals/corr/abs-2408-06142} &44.84  &43.78	&50.30	&55.97\\
jsl-medllama*-v18   &44.79  &42.83 &48.79  &43.39\\
TaxoLLaMA* \cite{DBLP:conf/acl/MoskvoretskiiNL24}       & 48.42                                   & 39.91                                   & 46.73                                   & 58.65                                   \\ \midrule
\textbf{OntoTune}$_{sft}$ & \textbf{53.02}                          & \underline{47.67}                          & \textbf{53.83}                          & \textbf{65.53} \\
\textbf{OntoTune}$_{dpo}$ & 50.46                          &\textbf{49.21}                        & \underline{53.61}                          &62.52                        \\
\textbf{OntoTune}$_{sft+dpo}$ &51.03                         & 45.22                          & 52.94                          & 62.81                         \\ \bottomrule
\end{tabular}}
\label{tab:hd}
\vspace{-2mm}
\end{table}

\subsection{Hypernym Discovery (RQ1)}
To verify whether the seed model can effectively align with the ontology, we evaluate the model's ontology reasoning ability through the in-ontology task hypernym discovery.
\subsubsection{\textbf{Datasets and Metric.}}
We select 4 subsets from the SemEval-2018 Task 9 \cite{DBLP:conf/semeval/Camacho-Collados18} dataset: 1A (English), 1B (Italian), 1C (Spanish), and 2A (Medical). The samples in these datasets contain a hyponym and a list of hypernyms, and the prompt template we used for training and evaluation is shown in Figure~\ref{fig:taxollama}. The performance is evaluated using the Mean Reciprocal Rank (MRR) metric denoted as $\operatorname{MRR}=\frac{1}{N} \sum_{i=1}^{N} \frac{1}{\operatorname{rank}_{i}}$, where $N$ is the total number of queries, and $rank_i$ is the rank of the correct result in the $i$-th query.

\begin{table*}[ht]
\centering
\caption{Results of the medical domain QA in the zero-shot and supervised fine-tuning (on evaluation) setting. The best results are highlighted in bold, while the second best are underlined. The TaxoLLaMA* represents the variants of TaxoLLaMA \cite{DBLP:conf/acl/MoskvoretskiiNL24} implemented by us. \textcolor{deepgreen}{$\uparrow$} and \textcolor{deepred}{$\downarrow$} indicate the score improvement and decline compared to the seed model.}
\vspace{-2mm}
\resizebox{0.98\textwidth}{!}{
\begin{tabular}{llllllllll}
\toprule \textbf{Setting}  & \textbf{Model}   & \textbf{MedQA} & \textbf{MedMCQA} & \textbf{PubMedQA} & \textbf{USMLE-step1} & \textbf{USMLE-step2} & \textbf{USMLE-step3} & \textbf{Average}  \\ \midrule
 \multirow{5}{*}{zero-shot} &LLaMA3 8B \cite{dubey2024llama}     & 51.7           &\quad 51.7             &\quad 70.3              &\quad \underline{57.4}                 &\quad 52.3                 &\quad 58.2                 & 56.9          \\
&TaxoLLaMA*\cite{DBLP:conf/acl/MoskvoretskiiNL24}     & 50.5\textcolor{deepred}{\;\small$\downarrow1.2$}           &\quad 46.1\textcolor{deepred}{\;\small$\downarrow5.6$}             &\quad \textbf{73.4}\textcolor{deepgreen}{\;\small$\uparrow3.1$}                &\quad 42.6\textcolor{deepred}{\;\small$\downarrow14.8$}                 &\quad 39.4\textcolor{deepred}{\;\small$\downarrow12.9$}                 &\quad 47.5\textcolor{deepred}{\;\small$\downarrow10.7$}                 & 49.9\textcolor{deepred}{\;\small$\downarrow7.0$}         \\  \cmidrule{2-10}
&\textbf{OntoTune$_{sft}$}   & 51.5\textcolor{deepred}{\;\small$\downarrow0.2$}           &\quad \underline{56.7}\textcolor{deepgreen}{\;\small$\uparrow5.0$}              &\quad \underline{72.0}\textcolor{deepgreen}{\;\small$\uparrow1.7$}                &\quad \underline{57.4}\textcolor{black}{\;-}                  &\quad \textbf{54.1}\textcolor{deepgreen}{\;\small$\uparrow1.8$}                   &\quad \underline{60.7}\textcolor{deepgreen}{\;\small$\uparrow2.5$}                   &\textbf{58.7}\textcolor{deepgreen}{\;\small$\uparrow1.8$}           \\
&\textbf{OntoTune$_{dpo}$}   & \textbf{53.3}\textcolor{deepgreen}{\;\small$\uparrow1.6$}           &\quad \textbf{57.2}\textcolor{deepgreen}{\;\small$\uparrow5.5$}             &\quad 65.5\textcolor{deepred}{\;\small$\downarrow4.8$}              &\quad \textbf{58.5}\textcolor{deepgreen}{\;\small$\uparrow1.1$}                 &\quad 51.4\textcolor{deepred}{\;\small$\downarrow0.9$}                 &\quad 59.0\textcolor{deepgreen}{\;\small$\uparrow0.8$}                 & 57.4\textcolor{deepgreen}{\;\small$\uparrow0.5$}         \\
&\textbf{OntoTune$_{sft+dpo}$}   & \underline{51.9}\textcolor{deepgreen}{\;\small$\uparrow0.2$}           &\quad \underline{56.7}\textcolor{deepgreen}{\;\small$\uparrow5.0$}             &\quad 66.3\textcolor{deepred}{\;\small$\downarrow4.0$}              &\quad 53.2\textcolor{deepred}{\;\small$\downarrow4.2$}                 &\quad \textbf{54.1}\textcolor{deepgreen}{\;\small$\uparrow1.8$}                 &\quad \textbf{63.1}\textcolor{deepgreen}{\;\small$\uparrow4.9$}                 & \underline{57.6}\textcolor{deepgreen}{\;\small$\uparrow0.7$}         \\ \midrule

\multirow{8}{*}{SFT (on evaluation)} &LLaMA3* 8B \cite{dubey2024llama}   & 56.4           &\quad 53.9             &\quad 77.2              &\quad 56.4                 &\quad 56.0                 &\quad 61.5                 & 60.2         \\
&Aloe \cite{DBLP:journals/corr/abs-2405-01886}    &53.4\textcolor{deepred}{\;\small$\downarrow3.0$}	&\quad56.8\textcolor{deepgreen}{\;\small$\uparrow2.9$}	&\quad75.4\textcolor{deepred}{\;\small$\downarrow1.8$}	&\quad54.3\textcolor{deepred}{\;\small$\downarrow2.1$}	&\quad\textbf{61.5}\textcolor{deepgreen}{\;\small$\uparrow5.5$}	&\quad60.7\textcolor{deepred}{\;\small$\downarrow0.8$}	&60.4\textcolor{deepgreen}{\;\small$\uparrow0.2$}   \\ 
&Med42-v2 \cite{DBLP:journals/corr/abs-2408-06142}    &57.8\textcolor{deepgreen}{\;\small$\uparrow1.4$}	&\quad58.1\textcolor{deepgreen}{\;\small$\uparrow4.2$}	&\quad74.6\textcolor{deepred}{\;\small$\downarrow2.6$}	&\quad\textbf{60.6}\textcolor{deepgreen}{\;\small$\uparrow4.2$}	&\quad\underline{57.8}\textcolor{deepgreen}{\;\small$\uparrow1.8$}	&\quad61.5\textcolor{black}{\;-}	&61.7\textcolor{deepgreen}{\;\small$\uparrow1.5$}   \\ 
&jsl-medllama-v18    &\textbf{59.3}\textcolor{deepgreen}{\;\small$\uparrow2.9$}	&\quad57.3\textcolor{deepgreen}{\;\small$\uparrow3.4$}	&\quad71.0\textcolor{deepred}{\;\small$\downarrow6.2$}	&\quad44.7\textcolor{deepred}{\;\small$\downarrow11.7$}	&\quad\underline{57.8}\textcolor{deepgreen}{\;\small$\uparrow1.8$}	&\quad\underline{62.3}\textcolor{deepgreen}{\;\small$\uparrow0.8$}		&58.7\textcolor{deepred}{\;\small$\downarrow1.5$}   \\ 
&TaxoLLaMA* \cite{DBLP:conf/acl/MoskvoretskiiNL24}   & 55.9\textcolor{deepred}{\;\small$\downarrow0.6$}           &\quad 57.5\textcolor{deepgreen}{\;\small$\uparrow3.6$}             &\quad 77.6\textcolor{deepgreen}{\;\small$\uparrow0.4$}              &\quad 56.4\textcolor{black}{\;-}                 &\quad \underline{57.8}\textcolor{deepgreen}{\;\small$\uparrow1.8$}                 &\quad 59.0\textcolor{deepred}{\;\small$\downarrow2.5$}                 & 60.7\textcolor{deepgreen}{\;\small$\uparrow0.5$}          \\ \cmidrule{2-10}
&\textbf{OntoTune}$_{sft}$     & \underline{58.4}\textcolor{deepgreen}{\;\small$\uparrow2.0$}  &\quad 60.4\textcolor{deepgreen}{\;\small$\uparrow6.5$}    &\quad 78.6\textcolor{deepgreen}{\;\small$\uparrow1.4$}     &\quad \underline{57.4}\textcolor{deepgreen}{\;\small$\uparrow1.0$}                 &\quad \underline{57.8}\textcolor{deepgreen}{\;\small$\uparrow1.8$}                 &\quad \underline{62.3}\textcolor{deepgreen}{\;\small$\uparrow0.8$}        & \textbf{62.5}\textcolor{deepgreen}{\;\small$\uparrow2.3$} \\ 
&\textbf{OntoTune}$_{dpo}$    & 58.3\textcolor{deepgreen}{\;\small$\uparrow1.9$}           &\quad \textbf{60.7}\textcolor{deepgreen}{\;\small$\uparrow6.8$}             &\quad \textbf{79.4}\textcolor{deepgreen}{\;\small$\uparrow2.2$}              &\quad 55.3\textcolor{deepred}{\;\small$\downarrow1.1$}                 &\quad 54.1\textcolor{deepred}{\;\small$\downarrow1.9$}                 &\quad 61.5\textcolor{black}{\;-}                 & 61.6\textcolor{deepgreen}{\;\small$\uparrow1.4$}         \\
&\textbf{OntoTune}$_{sft+dpo}$   & 58.2\textcolor{deepgreen}{\;\small$\uparrow1.8$}           &\quad \underline{60.5}\textcolor{deepgreen}{\;\small$\uparrow6.6$}             &\quad \underline{78.9}\textcolor{deepgreen}{\;\small$\uparrow2.2$}              &\quad \underline{57.4}\textcolor{deepgreen}{\;\small$\uparrow1.0$}                 &\quad 54.1\textcolor{deepred}{\;\small$\downarrow1.9$}                 &\quad \textbf{63.9}\textcolor{deepgreen}{\;\small$\uparrow2.4$}                 & \underline{62.2}\textcolor{deepgreen}{\;\small$\uparrow2.0$}    \\ \bottomrule
\end{tabular}}
\label{tab:qa}
\end{table*}

\subsubsection{\textbf{Baselines.}}
Our baselines can be divided into two part: \textbf{(1) embedding-based method}: CRIM \cite{DBLP:conf/semeval/Bernier-Colborne18}, Hybrid \cite{DBLP:conf/acl/HeldH19}, RMM \cite{DBLP:conf/acl/BaiZKCM21}, 300-sparsans \cite{DBLP:conf/semeval/BerendMF18}; \textbf{(2) PLM-based method}: T5$^*$ \cite{DBLP:conf/ijcnlp/NikishinaCDPB23}; \textbf{(3) LLM-based method}: LLaMA3 8B$^*$, TaxoLLaMA$^*$ \cite{DBLP:conf/acl/MoskvoretskiiNL24}, Aloe* \cite{DBLP:journals/corr/abs-2405-01886}, Med42-v2* \cite{DBLP:journals/corr/abs-2408-06142} and jsl-medllama*-3-8b-v18\footnote{https://huggingface.co/johnsnowlabs/jsl-medllama-3-8b-v18}. The T5$^*$ represents the taxonomy-adapted T5 \cite{DBLP:journals/jmlr/RaffelSRLNMZLL20} model implemented by Nikishina et al. \cite{DBLP:conf/ijcnlp/NikishinaCDPB23}. All LLM-based baselines and our OntoTune are developed based on LLaMA3 8B-Instruct, and have all been adapted for hypernym discovery task implemented by us. Among them, \textbf{TaxoLLaMA$^*$} \cite{DBLP:conf/acl/MoskvoretskiiNL24} is a direct ontology injection method. We adopt the same pre-training method as vanilla TaxoLLaMA \cite{DBLP:conf/acl/MoskvoretskiiNL24} and implement it with medical ontology SNOMED CT. Our instruction-tuning template is derived from the vanilla TaxoLLaMA \cite{DBLP:conf/acl/MoskvoretskiiNL24} as shown in Figure~\ref{fig:taxollama}, and it utilizes 510910 medical ontology relationships under the same training hyperparameters as OntoTune$_{sft}$. \textbf{Aloe*, Med42-v2*} and \textbf{jsl-medllama*-3-8b-v18} are medical LLMs fine-tuned on large-scale medical corpora and general instructions.
\subsubsection{\textbf{Implementation.}}
Considering the lack of definition of concepts in existing data sets \cite{DBLP:conf/coling/MoskvoretskiiPN24}, we follow previous generative work \cite{DBLP:conf/acl/MoskvoretskiiNL24} using GPT3.5-turbo\footnote{https://platform.openai.com/docs/models/gpt-3-5-turbo} to generate definitions for the hyponym concepts in these datasets, which helps to remove ambiguity. Additionally, we perform instruction-tuning for all LLM-based methods on the training set with a batch size of 32 per device and other training hyperparameters identical to OntoTune$_{sft}$.

\subsubsection{\textbf{Results.}}
\textit{\textbf{Medical Domain Performance.}} As shown in Table \ref{tab:hd}, the OntoTune$_{sft}$ models achieve state-of-the-art performance on the medical subset dataset, outperforming the seed model LLaMA* by 19.45\%, TaxoLLaMA* by 11.73\%. Although TaxoLLaMA* uses the entire SNOMED CT ontology for training, it does not achieve significant improvement. Moreover, we obverse that Aloe* and Med42-v2* trained on large-scale medical corpora exhibit noticeable performance improvements. Experimental results indicate that compared to TaxoLLaMA*, OntoTune can integrate ontology knowledge to LLMs more efficiently.

\textbf{\textit{Multilinual Performance.}} We conduct hypernym discovery tasks in multilingual environments, as shown in Table \ref{tab:hd}. Due to LLaMA3's pre-training in a multilingual environment, LLaMA* demonstrates good generalization performance on the Italian and Spanish subset datasets. However, TaxoLLaMA* and three medical LLMs experience catastrophic forgetting, with a significant performance decline compared to the seed model, whereas our three variants of OntoTune almost preserves the original multilingual capability of seed model. Notably, although our training set does not involve Italian and Spanish data, OntoTune$_{sft}$ also achieves state-of-the-art performance in the multilingual environment, showing significant improvement over seed model. This indicates that our OntoTune can effectively align seed model with ontology knowledge and even can generalize to other taxonomic scenarios.

\subsection{Medical Question Answering (RQ2)}
To verify whether seed model after being aligned with domain ontology, can effectively generalize to other domain-specific tasks, we conduct an out-of-ontology task domain QA for evaluation. 
\subsubsection{\textbf{Datasets.}}
We utilize 6 medical QA datasets: MedMCQA \cite{DBLP:conf/chil/PalUS22}, MedQA \cite{DBLP:journals/corr/abs-2009-13081}, PubMedQA \cite{DBLP:conf/emnlp/JinDLCL19}, USMLE step1-3 datasets\cite{DBLP:journals/corr/abs-2304-08247} to comprehensively evaluate medical domain ability. Among them, MedMCQA, MedQA, and PubMedQA have training sets. More details about the datasets can be found in Appendix \ref{app:a}.

\subsubsection{\textbf{Baselines.}}
To ensure a fair comparison, we only compare baselines based on the LLaMA3 8B-Instruct \cite{dubey2024llama}: \textbf{(1) existing domain LLM} based on large-scale corpora: Aloe \cite{DBLP:journals/corr/abs-2405-01886}, Med42-v2 \cite{DBLP:journals/corr/abs-2408-06142} and jsl-medllama-3-8b-v18; \textbf{(2) the direct ontology injection method TaxoLLaMA*} \cite{DBLP:conf/acl/MoskvoretskiiNL24}. We report the results for both zero-shot and supervised fine-tuning on the training set of the evaluation dataset. More baseline performances can be found in Appendix \ref{app:c}.

\begin{table*}[ht]
\centering
\caption{Results of general capabilities evaluation. \textcolor{deepgreen}{$\uparrow$} and \textcolor{deepred}{$\downarrow$} indicate the score improvement and decline of our OntoTune compared to the direct injection method TaxoLLaMA*.}
\resizebox{0.96\textwidth}{!}{
\begin{tabular}{llllll|ll|l|ll}
\toprule
\multirow{2}{*}{\textbf{Model}} & \multicolumn{5}{c}{\textbf{MMLU}}                                                              & \multicolumn{2}{c}{\textbf{ARC}}  & \multicolumn{1}{c}{\textbf{TriviaQA}} & \multicolumn{2}{c}{\textbf{Advbench}}       \\ \cmidrule{2-11}
                                & \textbf{STEM} & \textbf{Social Sciences} & \textbf{Humanities} & \textbf{Other} & \textbf{Average} & \textbf{ARC\_C} & \textbf{ARC\_E} & \qquad\;{\textbf{-}}         & \textbf{Raw Safe} & \textbf{Jailbreak Safe} \\ \midrule
LLaMA3  8B \cite{dubey2024llama}                     & 56.83         & \quad76.61                    & \quad60.81               & 74.10          & 66.49        & 78.64           & 92.77           & \,64.81             & \,97.50             & \quad96.35                   \\
Aloe \cite{DBLP:journals/corr/abs-2405-01886}	&55.67	&\quad\textbf{76.24}	&\quad58.91	&72.25	&65.10	&75.25	&86.95 &\,63.03  &\,62.50	&\quad34.23\\
Med42-v2 \cite{DBLP:journals/corr/abs-2408-06142}	&\textbf{56.59}	&\quad\textbf{76.24}	 &\quad59.91	&72.67	&65.72	&\textbf{82.37}	 &\textbf{92.59}	&\,\textbf{65.19}	&\,83.85	&\quad60.19 \\
jsl-medllama-v18	&55.07	&\quad74.13	&\quad58.00	&71.96	&64.13	&\underline{80.34}	&91.53	&\,61.59	&\,90.58	&\quad68.27 \\
\multicolumn{1}{l}{TaxoLLaMA* \cite{DBLP:conf/acl/MoskvoretskiiNL24}}  & 55.96         & \quad73.74                    & \quad56.92               & 69.43          & 63.29        & 72.88           & 89.24           & \,63.12             & \,\textbf{94.04}             & \quad73.27                   \\ \midrule
\textbf{OntoTune}$_{sft}$                        & \underline{56.47}\textcolor{deepgreen}{\;\small$\uparrow0.51$}         & \quad75.73 \textcolor{deepgreen}{\;\small$\uparrow1.99$}                    & \quad\textbf{61.85} \textcolor{deepgreen}{\;\small$\uparrow4.93$}               & \underline{73.02}\textcolor{deepgreen}{\;\small$\uparrow3.59$}           & \textbf{66.31} \textcolor{deepgreen}{\;\small$\uparrow3.02$}        & 78.31 \textcolor{deepgreen}{\;\small$\uparrow5.43$}           & 91.89\textcolor{deepgreen}{\;\small$\uparrow2.65$}            & \,\underline{64.07}\textcolor{deepgreen}{\;\small$\uparrow0.95$}              & \,\textbf{94.04}\textcolor{black}{\;\small$-$}              & \quad\textbf{92.69}\textcolor{deepgreen}{\;\small$\uparrow19.42$}                   \\ 
\textbf{OntoTune}$_{dpo}$                        & 56.33\textcolor{deepgreen}{\;\small$\uparrow0.37$}         & \quad75.33 \textcolor{deepgreen}{\;\small$\uparrow1.59$}                    & \quad59.93 \textcolor{deepgreen}{\;\small$\uparrow3.01$}               & \textbf{73.64}\textcolor{deepgreen}{\;\small$\uparrow4.21$}           & 65.70 \textcolor{deepgreen}{\;\small$\uparrow2.41$}        & 78.98 \textcolor{deepgreen}{\;\small$\uparrow6.10$}           & \underline{92.06}\textcolor{deepgreen}{\;\small$\uparrow2.82$}            & \,63.96\textcolor{deepgreen}{\;\small$\uparrow0.84$}              & \,\underline{90.58}\textcolor{deepred}{\;\small$\downarrow3.46$}              & \quad77.88\textcolor{deepgreen}{\;\small$\uparrow4.61$}                   \\ 
\textbf{OntoTune}$_{sft+dpo}$                        & 55.67\textcolor{deepred}{\;\small$\downarrow0.29$}         & \quad75.17 \textcolor{deepgreen}{\;\small$\uparrow1.43$}                    & \quad\underline{61.79} \textcolor{deepgreen}{\;\small$\uparrow4.87$}               & 72.71\textcolor{deepgreen}{\;\small$\uparrow3.28$}           & \underline{65.93} \textcolor{deepgreen}{\;\small$\uparrow2.64$}        & 78.98 \textcolor{deepgreen}{\;\small$\uparrow6.10$}           & \underline{92.06}\textcolor{deepgreen}{\;\small$\uparrow2.82$}            & \,63.96\textcolor{deepgreen}{\;\small$\uparrow0.84$}                & \,\underline{90.58}\textcolor{deepred}{\;\small$\downarrow3.46$}              & \quad\underline{84.81}\textcolor{deepgreen}{\;\small$\uparrow11.54$}                   \\ \bottomrule
\end{tabular}}
\label{tab:gc}
\end{table*}

\subsubsection{\textbf{Implementation.}}
Following previous works \cite{DBLP:journals/corr/abs-2304-08247,DBLP:conf/acl/LabrakBMGRD24,DBLP:journals/corr/abs-2404-16621}, we perform instruction-tuning on the training set of the evaluation dataset for LLaMA3, TaxoLLaMA and OntoTune with the same training hyperparameters as OntoTune$_{sft}$.

\subsubsection{\textbf{Results.}}
From zero-shot results shown in Table \ref{tab:qa}, we can observe that the performance of TaxoLLaMA* significantly declines and the performance of OntoTune increases on most datasets. And when we conduct supervised fine-tuning on the instruction dataset, OntoTune$_{sft}$ performs better than seed model across all datasets and achieves state-of-the-art results among all LLMs based on LLaMA3 8B. Compared to our seed model, all three variants of our OntoTune, as well as the TaxoLLaMA* method, achieve significant improvements. This indicates that a small-scale, but high-quality ontology is beneficial for enhancing the capabilities of LLMs in specific domains. It's observed that although LLMs trained on large-scale raw corpora perform well on some datasets, their improvement over the seed model is not stable and the average score is inferior to our OntoTune, which suggests that large-scale corpora are challenging to learn from. To our surprise, although ontologies cannot directly provide the concrete knowledge related to these practical questions for the seed model, we attribute the performance improvement to the structured ontology knowledge, which helps LLMs reorganize domain knowledge. Furthermore, our three OntoTune models outperform the direct ontology injection method TaxoLLaMA*, demonstrating self-training is more effective for reorganizing domain knowledge and improving the performance of domain-specific tasks.

\begin{figure}[t]
\centering
\includegraphics[scale=0.415]{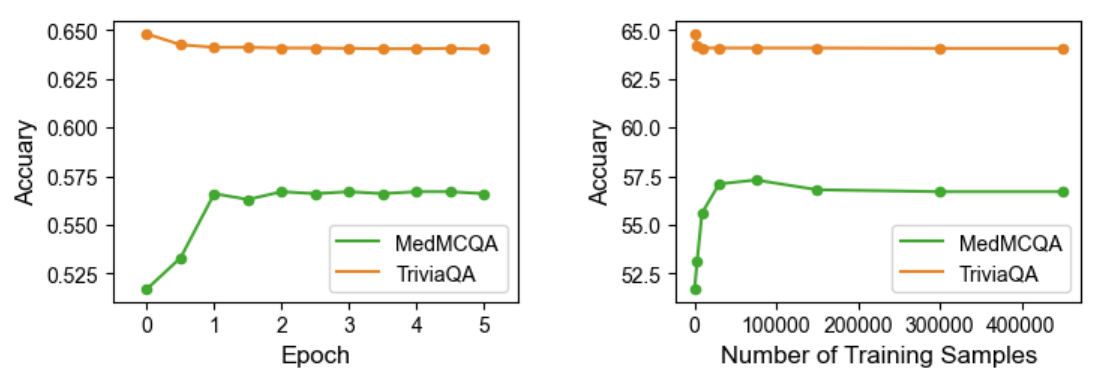}
\vspace{-3.5mm}
\caption{Performance with different epochs and training samples. The result of MedMCQA is under zero-shot setting.}
\label{fig:epoch}
\end{figure}

\subsection{General Capabilities Evaluation (RQ3)}
Futhermore, we evaluate whether the seed model exhibits catastrophic forgetting or impaired capabilities after OntoTune.
\subsubsection{\textbf{Knowledge Evaluation.}}
We conduct evaluation on the MMLU \cite{DBLP:conf/iclr/HendrycksBBZMSS21}, ARC \cite{DBLP:journals/corr/abs-1803-05457}, and TrivialQA \cite{DBLP:conf/acl/JoshiCWZ17} datasets. Specifically, MMLU is evaluated based on LLaMA-Factory \cite{zheng2024llamafactory}, while ARC and TrivialQA are evaluated on OpenCompass \cite{2023opencompass} tool with gen mode.

From the results in Table \ref{tab:gc}, we observe that Med42-v2 even surpasses the seed model on several datasets. This is because Med42-v2 incorporates 344k general instructions during the domain adaptation phase, with 74k CoT instructions effectively enhancing reasoning performance on the ARC dataset. In contrast, other domain LLMs that also incorporate general instructions experience a noticeable decline in general performance compared to our OntoTune, which does not use general instructions. Additionally, due to the fixed input-output format and lack of data diversity \cite{DBLP:conf/nips/ZhouLX0SMMEYYZG23}, TaxoLLaMA* suffers the most significant performance decline. Compared to TaxoLLaMA*, our OntoTune method does not exhibit significant catastrophic forgetting. Similarly, OntoTune$_{sft}$ demonstrates the best performance among three variants, showing an average decrease of only 0.49\% compared to the seed model.

\subsubsection{\textbf{Safety Evaluation.}}
Following previous work \cite{DBLP:conf/iclr/Qi0XC0M024,DBLP:conf/acl/YangPFWCZL24} on safety evaluation, we select harmful instructions from the Advbench dataset \cite{DBLP:journals/corr/abs-2307-15043} as model inputs and denote the proportion of safe responses as ``Raw Safe". Then we append adversarial suffixes to the harmful instructions and denote the proportion of safe responses at present as ``Jailbreak Safe" to measure model's safety.

\begin{figure}[t]
\centering
\includegraphics[scale=0.323]{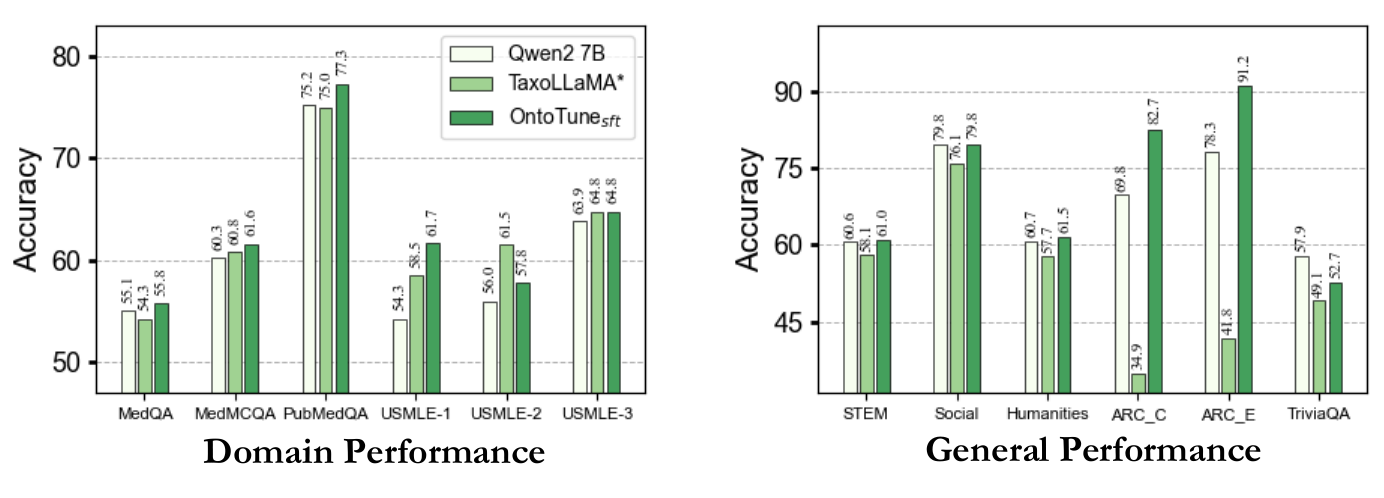}
\vspace{-4mm}
\caption{Domain performances and general performances on the seed model Qwen2 7B.}
\vspace{-1mm}
\label{fig:qwen}

\end{figure}

From results in Table \ref{tab:gc}, we observe that the fine-tuned models show a significant decline in both Raw Safe and Jailbreak Safe metrics. Despite undergoing safety alignment, the three medical models based on large-scale corpora still exhibit catastrophic security vulnerabilities. For four ontology-based fine-tuning approach, TaxoLLaMA* and OntoTune both show a slight decline in the Raw Safe metric. Under jailbreak settings, TaxoLLaMA* experiences a significant 23.08\% decline in the Jailbreak Safe metric, while OntoTune effectively mitigates this issue. OntoTune demonstrates state-of-the-art performance, not only achieving efficient domain alignment but also preserving safety alignment.

\subsection{\textbf{Model Analysis}}

\subsubsection{\textbf{Effects of Training Parameters.}}

In Figure \ref{fig:epoch}, we explore the performance of our OntoTune across different training epochs and different numbers of samples. Specifically, we use TriviaQA to evaluate general performance and MedMCQA to evaluate domain-specific performance. We find that with 300,000 training samples, just 1 epoch leads to significant performance improvement. Additionally, at 3 training epochs, there is a noticeable improvement with only 9,000 samples, and the seed model trained on 75,000 samples achieves best performance. As the amount of training and data volume increase, OntoTune gradually converges. This implies that compared to existing domain LLMs, we can achieve more robust results using fewer training samples through OntoTune.

\subsubsection{\textbf{Robustness to Seed Models.}}
We use Qwen2 7B \cite{DBLP:journals/corr/abs-2407-10671} as the seed model and report the performance of TaxoLLaMA* and the best variant, OntoTune$_{sft}$ to demonstrate that OntoTune is not

\noindent constrained by model architecture. As shown in Figure \ref{fig:qwen}, OntoTune$_{sft}$ achieves improvements over the base model across all medical QA datasets. Notably, OntoTune$_{sft}$ even achieves improvements on most of the general datasets, and significantly enhances reasoning performance on ARC. This improvement may be due to the enhancement of planning abilities when trained with structured data \cite{DBLP:journals/corr/abs-2406-14282}. Conversely, although TaxoLLaMA* shows improvement in medical QA, it experiences a significant decline in general performance. These results suggest that aligning with ontology benefits domain-specific capabilities, demonstrating OntoTune's robustness.
\begin{table}
\caption{Results of domain capabilities for the three variants of OntoTune$_{sft}$ on LLaMA3 8B. The reference outputs $y^o$ in their training sets are from self-generated by LLaMA3 8B, and distilled by LLaMA3.1 8B and deepseek-chat. 
}
\vspace{-2mm}
\centering
\resizebox{0.46\textwidth}{!}{
\begin{tabular}{cccccc}
\toprule
$y^o$ \textbf{from}       & \textbf{MedQA} & \textbf{MedMCQA} & \textbf{PubMedQA} & \textbf{USMLE-Avg} & \textbf{Average}  \\ \midrule
 -             & 56.4  & 53.9    & 77.2     & \underline{58.0}        & 60.2 \\
LLaMA3 8B     & \textbf{58.4}  & \underline{60.4}    & \textbf{78.6}     & \textbf{59.2}        & \textbf{62.5} \\
 LLaMA3.1 8B   & \textbf{58.4}  & \textbf{61.1}    & \underline{77.7}     & 56.1       & 60.9 \\
deepseek-chat & 57.8  & 60.2    & 77.2     & 57.2        & \underline{61.2} \\ \bottomrule
\end{tabular}}
\label{tab:distill_qa}
\vspace{-1mm}
\end{table}

\begin{figure}
\centering
\includegraphics[scale=0.37]{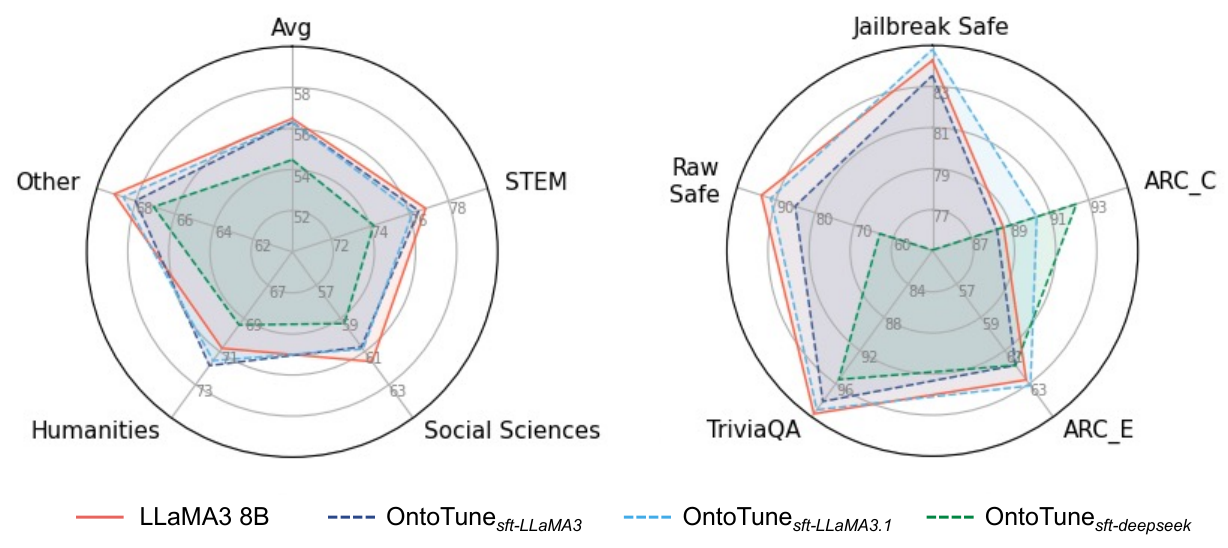}
\vspace{-1mm}
\caption{General performances for the three variants of OntoTune$_{sft}$ on LLaMA3 8B.}
\vspace{-2mm}
\label{fig:self-training}
\end{figure}

\subsubsection{\textbf{Self-training Analysis.}}
Aiming to explore the impact of data quality on model's performance, we distill two stronger LLMs: LLaMA 3.1 8B and deepseek-v2.5-chat\footnote{https://chat.deepseek.com/}, using $x^o_t=\{(x,t,o_t)\,|\,y^o_t=f_\theta(x,t,o_t),y^o_t\in\mathcal{D}_{sft}\}$ as input to generate the higher quality target output $y^{o'}$. We then train the same seed model on $\mathcal{D}_{sft}'=\{x_n,y_n^{o'}\}_{n=1}^k$ under the same hyperparameters settings. Table \ref{tab:distill_qa} presents the results of three LLMs compared to the seed model in domain QA. On most datasets, the performances of all three variants of OntoTune can be improved. Among them, the self-training OntoTune$_{sft}$ model demonstrates robust and advanced performance, achieving improvements across all datasets. From results in Figure \ref{fig:self-training}, We observe that the OntoTune$_{sft}$ distilled from the same series LLaMA 3.1, exhibits the least decline on the knowledge QA dataset like MMLU and TriviaQA. Interestingly, although the focus is only on medical domain knowledge during the data distillation of LLaMA 3.1, the model shows improved performance on the reasoning challenge dataset ARC and safety evaluation Advbench. Additionally, the model distilled from deepseek shows a noticeable decline in knowledge and safety evaluation but a significant enhancement in reasoning ability. Overall, self-training achieves the most efficient domain alignment without requiring

\noindent advanced LLMs, while greatly preserving original knowledge.

\subsubsection{\textbf{Distribution Shift Analysis.}}
In the preceding sections, we identify OntoTune$_{sft}$ as the variant with best performance, excelling not only in downstream tasks but also effectively preserving the knowledge and safety of the seed model. We attribute this phenomenon to distribution shift. We utilize the mean squared change in parameters (denoted as $|\Delta\theta|^2$) to measure parameter shift during training and evaluate the data distribution shift based on the similarity of the model's responses. Specifically, we collect 1,000 general instructions from the Alpaca evaluation set \cite{alpaca_eval} and use the seed model's responses to these instructions as reference responses. We calculate the cosine similarity between the fine-tuned model's responses and the reference responses. 

\begin{figure}[t]
\centering
\includegraphics[scale=0.38]{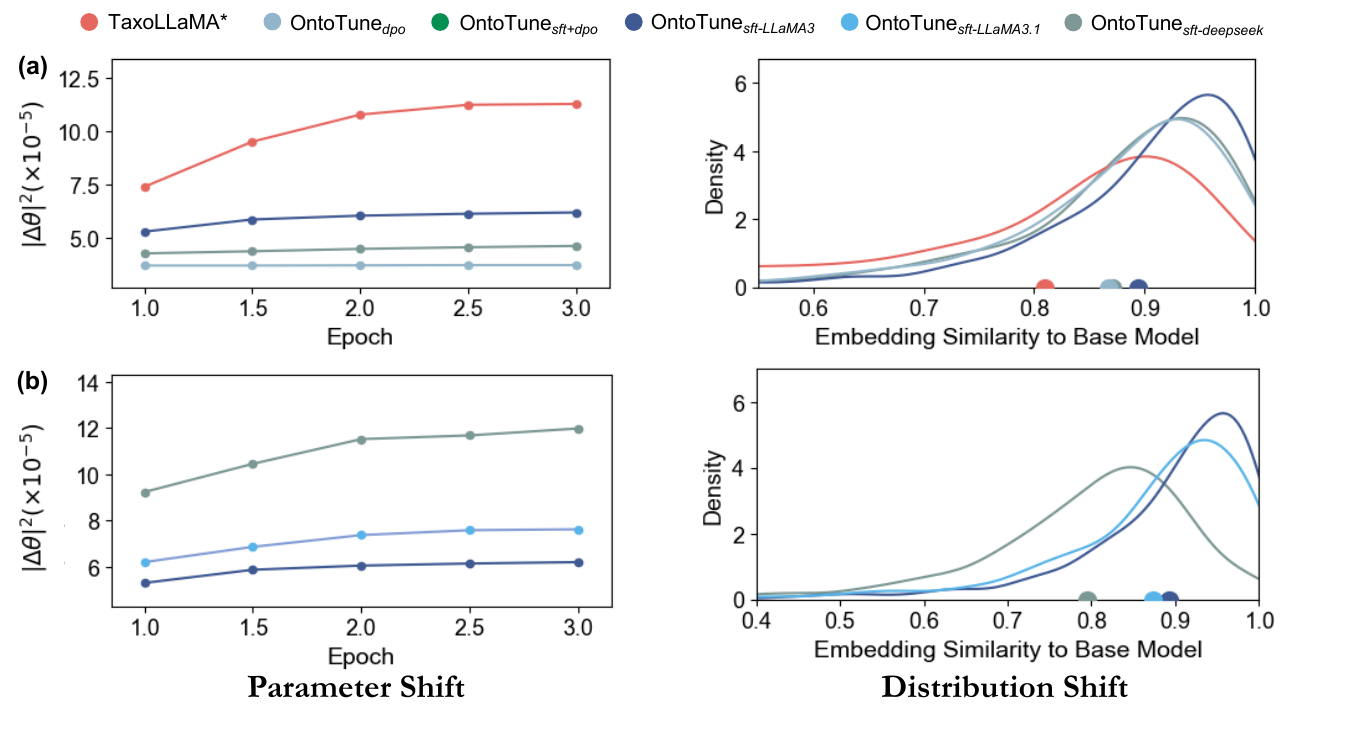}
\vspace{-7mm}
\caption{(a) Comparison of OntoTune variants and TaxoLLaMA*. (b) Comparison of data distillation and self-training.}
\vspace{-2mm}
\label{fig:shift}
\end{figure}

From results shown in Figure \ref{fig:shift}, it can be observed that OntoTune$_{sft}$ exhibits the largest parameter shift, but it exhibits the least data distribution shift. Compared to distilling a larger LLM, the parameter and data distribution shifts in the self-training setting are smaller. Additionally, distilling from the same series LLM results in less distribution shift, which we infer is due to the similar pre-training data. Therefore, we can obtain the conclusion consistent with previous research \cite{DBLP:conf/acl/YangPFWCZL24}: self-training can effectively bridge distribution gap and thereby mitigate catastrophic forgetting.

\section{Conclusion}
In this paper, we propose an ontology-driven self-training fine-tuning framework \textbf{OntoTune},
which leverages in-context learning to identify the specific concept’s ontology knowledge the seed model has not acquired, and perform self-training to enhance the seed model's alignment with the ontology. Experiments demonstrate that OntoTune achieves state-of-the-art performance in both in-ontology task hypernym discovery and out-of-ontology task medical domain QA, while significantly preserving the knowledge of the seed model. Compared to existing domain LLMs trained on large-scale high-quality corpora, OntoTune relies on a relatively small-scale, long-term developed ontology along with the seed model itself, offering improved generalization ability. 
In the future, we will explore automated alignment methods that are less dependent on specific instruction templates. And we hope OntoTune could inspire more researches into exploring more efficient domain adaptation methods using small-scale data when facing the rapid iteration of LLMs and the scarcity of domain-specific data.  
\begin{acks}
This work is founded by National Natural Science Foundation of China (NSFC62306276/NSFCU23B2055/NSFCU19B2027), Zhejiang Provincial Natural Science Foundation of China (No. LQ23F020017), Yongjiang Talent Introduction Programme (2022A-238-G), and Fundamental Research Funds for the Central Universities (226-2023-00138). This work was supported by AntGroup.
\end{acks}


\bibliographystyle{ACM-Reference-Format}
\balance
\bibliography{ref}


\begin{thebibliography}{68}


\ifx \showCODEN    \undefined \def \showCODEN     #1{\unskip}     \fi
\ifx \showISBNx    \undefined \def \showISBNx     #1{\unskip}     \fi
\ifx \showISBNxiii \undefined \def \showISBNxiii  #1{\unskip}     \fi
\ifx \showISSN     \undefined \def \showISSN      #1{\unskip}     \fi
\ifx \showLCCN     \undefined \def \showLCCN      #1{\unskip}     \fi
\ifx \shownote     \undefined \def \shownote      #1{#1}          \fi
\ifx \showarticletitle \undefined \def \showarticletitle #1{#1}   \fi
\ifx \showURL      \undefined \def \showURL       {\relax}        \fi
\providecommand\bibfield[2]{#2}
\providecommand\bibinfo[2]{#2}
\providecommand\natexlab[1]{#1}
\providecommand\showeprint[2][]{arXiv:#2}

\bibitem[Acikgoz et~al\mbox{.}(2024)]%
        {DBLP:journals/corr/abs-2404-16621}
\bibfield{author}{\bibinfo{person}{Emre~Can Acikgoz}, \bibinfo{person}{Osman~Batur Ince}, \bibinfo{person}{Rayene Bench}, \bibinfo{person}{Arda~Anil Boz}, \bibinfo{person}{Ilker Kesen}, \bibinfo{person}{Aykut Erdem}, {and} \bibinfo{person}{Erkut Erdem}.} \bibinfo{year}{2024}\natexlab{}.
\newblock \showarticletitle{Hippocrates: An Open-Source Framework for Advancing Large Language Models in Healthcare}.
\newblock \bibinfo{journal}{\emph{CoRR}}  \bibinfo{volume}{abs/2404.16621} (\bibinfo{year}{2024}).
\newblock
\href{https://doi.org/10.48550/ARXIV.2404.16621}{doi:\nolinkurl{10.48550/ARXIV.2404.16621}}
\showeprint[arXiv]{2404.16621}


\bibitem[Almeida et~al\mbox{.}(2024)]%
        {DBLP:journals/ai/AlmeidaNEWA24}
\bibfield{author}{\bibinfo{person}{Guilherme F. C.~F. Almeida}, \bibinfo{person}{Jos{\'{e}}~Luiz Nunes}, \bibinfo{person}{Neele Engelmann}, \bibinfo{person}{Alex Wiegmann}, {and} \bibinfo{person}{Marcelo de Ara{\'{u}}jo}.} \bibinfo{year}{2024}\natexlab{}.
\newblock \showarticletitle{Exploring the psychology of LLMs' moral and legal reasoning}.
\newblock \bibinfo{journal}{\emph{Artif. Intell.}}  \bibinfo{volume}{333} (\bibinfo{year}{2024}), \bibinfo{pages}{104145}.
\newblock
\href{https://doi.org/10.1016/J.ARTINT.2024.104145}{doi:\nolinkurl{10.1016/J.ARTINT.2024.104145}}


\bibitem[Amini et~al\mbox{.}(2022)]%
        {DBLP:journals/corr/abs-2202-12040}
\bibfield{author}{\bibinfo{person}{Massih{-}Reza Amini}, \bibinfo{person}{Vasilii Feofanov}, \bibinfo{person}{Lo{\"{\i}}c Pauletto}, \bibinfo{person}{Emilie Devijver}, {and} \bibinfo{person}{Yury Maximov}.} \bibinfo{year}{2022}\natexlab{}.
\newblock \showarticletitle{Self-Training: {A} Survey}.
\newblock \bibinfo{journal}{\emph{CoRR}}  \bibinfo{volume}{abs/2202.12040} (\bibinfo{year}{2022}).
\newblock
\showeprint[arXiv]{2202.12040}
\urldef\tempurl%
\url{https://arxiv.org/abs/2202.12040}
\showURL{%
\tempurl}


\bibitem[Bai et~al\mbox{.}(2021)]%
        {DBLP:conf/acl/BaiZKCM21}
\bibfield{author}{\bibinfo{person}{Yuhang Bai}, \bibinfo{person}{Richong Zhang}, \bibinfo{person}{Fanshuang Kong}, \bibinfo{person}{Junfan Chen}, {and} \bibinfo{person}{Yongyi Mao}.} \bibinfo{year}{2021}\natexlab{}.
\newblock \showarticletitle{Hypernym Discovery via a Recurrent Mapping Model}. In \bibinfo{booktitle}{\emph{Findings of the Association for Computational Linguistics: {ACL/IJCNLP} 2021, Online Event, August 1-6, 2021}} \emph{(\bibinfo{series}{Findings of {ACL}}, Vol.~\bibinfo{volume}{{ACL/IJCNLP} 2021})}, \bibfield{editor}{\bibinfo{person}{Chengqing Zong}, \bibinfo{person}{Fei Xia}, \bibinfo{person}{Wenjie Li}, {and} \bibinfo{person}{Roberto Navigli}} (Eds.). \bibinfo{publisher}{Association for Computational Linguistics}, \bibinfo{pages}{2912--2921}.
\newblock
\href{https://doi.org/10.18653/V1/2021.FINDINGS-ACL.257}{doi:\nolinkurl{10.18653/V1/2021.FINDINGS-ACL.257}}


\bibitem[Berend et~al\mbox{.}(2018)]%
        {DBLP:conf/semeval/BerendMF18}
\bibfield{author}{\bibinfo{person}{G{\'{a}}bor Berend}, \bibinfo{person}{M{\'{a}}rton Makrai}, {and} \bibinfo{person}{Peter F{\"{o}}ldi{\'{a}}k}.} \bibinfo{year}{2018}\natexlab{}.
\newblock \showarticletitle{300-sparsans at SemEval-2018 Task 9: Hypernymy as interaction of sparse attributes}. In \bibinfo{booktitle}{\emph{Proceedings of The 12th International Workshop on Semantic Evaluation, SemEval@NAACL-HLT 2018, New Orleans, Louisiana, USA, June 5-6, 2018}}, \bibfield{editor}{\bibinfo{person}{Marianna Apidianaki}, \bibinfo{person}{Saif~M. Mohammad}, \bibinfo{person}{Jonathan May}, \bibinfo{person}{Ekaterina Shutova}, \bibinfo{person}{Steven Bethard}, {and} \bibinfo{person}{Marine Carpuat}} (Eds.). \bibinfo{publisher}{Association for Computational Linguistics}, \bibinfo{pages}{928--934}.
\newblock
\href{https://doi.org/10.18653/V1/S18-1152}{doi:\nolinkurl{10.18653/V1/S18-1152}}


\bibitem[Bernier{-}Colborne and Barri{\`{e}}re(2018)]%
        {DBLP:conf/semeval/Bernier-Colborne18}
\bibfield{author}{\bibinfo{person}{Gabriel Bernier{-}Colborne} {and} \bibinfo{person}{Caroline Barri{\`{e}}re}.} \bibinfo{year}{2018}\natexlab{}.
\newblock \showarticletitle{{CRIM} at SemEval-2018 Task 9: {A} Hybrid Approach to Hypernym Discovery}. In \bibinfo{booktitle}{\emph{Proceedings of The 12th International Workshop on Semantic Evaluation, SemEval@NAACL-HLT 2018, New Orleans, Louisiana, USA, June 5-6, 2018}}, \bibfield{editor}{\bibinfo{person}{Marianna Apidianaki}, \bibinfo{person}{Saif~M. Mohammad}, \bibinfo{person}{Jonathan May}, \bibinfo{person}{Ekaterina Shutova}, \bibinfo{person}{Steven Bethard}, {and} \bibinfo{person}{Marine Carpuat}} (Eds.). \bibinfo{publisher}{Association for Computational Linguistics}, \bibinfo{pages}{725--731}.
\newblock
\href{https://doi.org/10.18653/V1/S18-1116}{doi:\nolinkurl{10.18653/V1/S18-1116}}


\bibitem[Bhatia et~al\mbox{.}(2024)]%
        {DBLP:conf/acl/BhatiaNCA24}
\bibfield{author}{\bibinfo{person}{Gagan Bhatia}, \bibinfo{person}{El~Moatez~Billah Nagoudi}, \bibinfo{person}{Hasan Cavusoglu}, {and} \bibinfo{person}{Muhammad Abdul{-}Mageed}.} \bibinfo{year}{2024}\natexlab{}.
\newblock \showarticletitle{FinTral: {A} Family of {GPT-4} Level Multimodal Financial Large Language Models}. In \bibinfo{booktitle}{\emph{Findings of the Association for Computational Linguistics, {ACL} 2024, Bangkok, Thailand and virtual meeting, August 11-16, 2024}}, \bibfield{editor}{\bibinfo{person}{Lun{-}Wei Ku}, \bibinfo{person}{Andre Martins}, {and} \bibinfo{person}{Vivek Srikumar}} (Eds.). \bibinfo{publisher}{Association for Computational Linguistics}, \bibinfo{pages}{13064--13087}.
\newblock
\urldef\tempurl%
\url{https://aclanthology.org/2024.findings-acl.774}
\showURL{%
\tempurl}


\bibitem[Camacho{-}Collados et~al\mbox{.}(2018)]%
        {DBLP:conf/semeval/Camacho-Collados18}
\bibfield{author}{\bibinfo{person}{Jos{\'{e}} Camacho{-}Collados}, \bibinfo{person}{Claudio~Delli Bovi}, \bibinfo{person}{Luis~Espinosa Anke}, \bibinfo{person}{Sergio Oramas}, \bibinfo{person}{Tommaso Pasini}, \bibinfo{person}{Enrico Santus}, \bibinfo{person}{Vered Shwartz}, \bibinfo{person}{Roberto Navigli}, {and} \bibinfo{person}{Horacio Saggion}.} \bibinfo{year}{2018}\natexlab{}.
\newblock \showarticletitle{SemEval-2018 Task 9: Hypernym Discovery}. In \bibinfo{booktitle}{\emph{Proceedings of The 12th International Workshop on Semantic Evaluation, SemEval@NAACL-HLT 2018, New Orleans, Louisiana, USA, June 5-6, 2018}}, \bibfield{editor}{\bibinfo{person}{Marianna Apidianaki}, \bibinfo{person}{Saif~M. Mohammad}, \bibinfo{person}{Jonathan May}, \bibinfo{person}{Ekaterina Shutova}, \bibinfo{person}{Steven Bethard}, {and} \bibinfo{person}{Marine Carpuat}} (Eds.). \bibinfo{publisher}{Association for Computational Linguistics}, \bibinfo{pages}{712--724}.
\newblock
\href{https://doi.org/10.18653/V1/S18-1115}{doi:\nolinkurl{10.18653/V1/S18-1115}}


\bibitem[Chen et~al\mbox{.}(2023)]%
        {DBLP:conf/icde/00070HCGYBZYSWY23}
\bibfield{author}{\bibinfo{person}{Zhuo Chen}, \bibinfo{person}{Wen Zhang}, \bibinfo{person}{Yufeng Huang}, \bibinfo{person}{Mingyang Chen}, \bibinfo{person}{Yuxia Geng}, \bibinfo{person}{Hongtao Yu}, \bibinfo{person}{Zhen Bi}, \bibinfo{person}{Yichi Zhang}, \bibinfo{person}{Zhen Yao}, \bibinfo{person}{Wenting Song}, \bibinfo{person}{Xinliang Wu}, \bibinfo{person}{Yi Yang}, \bibinfo{person}{Mingyi Chen}, \bibinfo{person}{Zhaoyang Lian}, \bibinfo{person}{Yingying Li}, \bibinfo{person}{Lei Cheng}, {and} \bibinfo{person}{Huajun Chen}.} \bibinfo{year}{2023}\natexlab{}.
\newblock \showarticletitle{Tele-Knowledge Pre-training for Fault Analysis}. In \bibinfo{booktitle}{\emph{39th {IEEE} International Conference on Data Engineering, {ICDE} 2023, Anaheim, CA, USA, April 3-7, 2023}}. \bibinfo{publisher}{{IEEE}}, \bibinfo{pages}{3453--3466}.
\newblock
\href{https://doi.org/10.1109/ICDE55515.2023.00265}{doi:\nolinkurl{10.1109/ICDE55515.2023.00265}}


\bibitem[Cheng et~al\mbox{.}(2024)]%
        {DBLP:conf/iclr/ChengHW24}
\bibfield{author}{\bibinfo{person}{Daixuan Cheng}, \bibinfo{person}{Shaohan Huang}, {and} \bibinfo{person}{Furu Wei}.} \bibinfo{year}{2024}\natexlab{}.
\newblock \showarticletitle{Adapting Large Language Models via Reading Comprehension}. In \bibinfo{booktitle}{\emph{The Twelfth International Conference on Learning Representations, {ICLR} 2024, Vienna, Austria, May 7-11, 2024}}. \bibinfo{publisher}{OpenReview.net}.
\newblock
\urldef\tempurl%
\url{https://openreview.net/forum?id=y886UXPEZ0}
\showURL{%
\tempurl}


\bibitem[Christophe et~al\mbox{.}(2024)]%
        {DBLP:journals/corr/abs-2408-06142}
\bibfield{author}{\bibinfo{person}{Cl{\'{e}}ment Christophe}, \bibinfo{person}{Praveen~K. Kanithi}, \bibinfo{person}{Tathagata Raha}, \bibinfo{person}{Shadab Khan}, {and} \bibinfo{person}{Marco~AF Pimentel}.} \bibinfo{year}{2024}\natexlab{}.
\newblock \showarticletitle{Med42-v2: {A} Suite of Clinical LLMs}.
\newblock \bibinfo{journal}{\emph{CoRR}}  \bibinfo{volume}{abs/2408.06142} (\bibinfo{year}{2024}).
\newblock
\href{https://doi.org/10.48550/ARXIV.2408.06142}{doi:\nolinkurl{10.48550/ARXIV.2408.06142}}
\showeprint[arXiv]{2408.06142}


\bibitem[Clark et~al\mbox{.}(2018)]%
        {DBLP:journals/corr/abs-1803-05457}
\bibfield{author}{\bibinfo{person}{Peter Clark}, \bibinfo{person}{Isaac Cowhey}, \bibinfo{person}{Oren Etzioni}, \bibinfo{person}{Tushar Khot}, \bibinfo{person}{Ashish Sabharwal}, \bibinfo{person}{Carissa Schoenick}, {and} \bibinfo{person}{Oyvind Tafjord}.} \bibinfo{year}{2018}\natexlab{}.
\newblock \showarticletitle{Think you have Solved Question Answering? Try ARC, the {AI2} Reasoning Challenge}.
\newblock \bibinfo{journal}{\emph{CoRR}}  \bibinfo{volume}{abs/1803.05457} (\bibinfo{year}{2018}).
\newblock
\showeprint[arXiv]{1803.05457}
\urldef\tempurl%
\url{http://arxiv.org/abs/1803.05457}
\showURL{%
\tempurl}


\bibitem[Contributors(2023)]%
        {2023opencompass}
\bibfield{author}{\bibinfo{person}{OpenCompass Contributors}.} \bibinfo{year}{2023}\natexlab{}.
\newblock \bibinfo{title}{OpenCompass: A Universal Evaluation Platform for Foundation Models}.
\newblock \bibinfo{howpublished}{\url{https://github.com/open-compass/opencompass}}.
\newblock


\bibitem[Dorfner et~al\mbox{.}(2024)]%
        {dorfner2024biomedical}
\bibfield{author}{\bibinfo{person}{Felix~J Dorfner}, \bibinfo{person}{Amin Dada}, \bibinfo{person}{Felix Busch}, \bibinfo{person}{Marcus~R Makowski}, \bibinfo{person}{Tianyu Han}, \bibinfo{person}{Daniel Truhn}, \bibinfo{person}{Jens Kleesiek}, \bibinfo{person}{Madhumita Sushil}, \bibinfo{person}{Jacqueline Lammert}, \bibinfo{person}{Lisa~C Adams}, {et~al\mbox{.}}} \bibinfo{year}{2024}\natexlab{}.
\newblock \showarticletitle{Biomedical Large Languages Models Seem not to be Superior to Generalist Models on Unseen Medical Data}.
\newblock \bibinfo{journal}{\emph{arXiv preprint arXiv:2408.13833}} (\bibinfo{year}{2024}).
\newblock


\bibitem[Dubey et~al\mbox{.}(2024)]%
        {dubey2024llama}
\bibfield{author}{\bibinfo{person}{Abhimanyu Dubey}, \bibinfo{person}{Abhinav Jauhri}, \bibinfo{person}{Abhinav Pandey}, \bibinfo{person}{Abhishek Kadian}, \bibinfo{person}{Ahmad Al-Dahle}, \bibinfo{person}{Aiesha Letman}, \bibinfo{person}{Akhil Mathur}, \bibinfo{person}{Alan Schelten}, \bibinfo{person}{Amy Yang}, \bibinfo{person}{Angela Fan}, {et~al\mbox{.}}} \bibinfo{year}{2024}\natexlab{}.
\newblock \showarticletitle{The llama 3 herd of models}.
\newblock \bibinfo{journal}{\emph{arXiv preprint arXiv:2407.21783}} (\bibinfo{year}{2024}).
\newblock


\bibitem[Gekhman et~al\mbox{.}(2024)]%
        {DBLP:journals/corr/abs-2405-05904}
\bibfield{author}{\bibinfo{person}{Zorik Gekhman}, \bibinfo{person}{Gal Yona}, \bibinfo{person}{Roee Aharoni}, \bibinfo{person}{Matan Eyal}, \bibinfo{person}{Amir Feder}, \bibinfo{person}{Roi Reichart}, {and} \bibinfo{person}{Jonathan Herzig}.} \bibinfo{year}{2024}\natexlab{}.
\newblock \showarticletitle{Does Fine-Tuning LLMs on New Knowledge Encourage Hallucinations?}
\newblock \bibinfo{journal}{\emph{CoRR}}  \bibinfo{volume}{abs/2405.05904} (\bibinfo{year}{2024}).
\newblock
\href{https://doi.org/10.48550/ARXIV.2405.05904}{doi:\nolinkurl{10.48550/ARXIV.2405.05904}}
\showeprint[arXiv]{2405.05904}


\bibitem[Gururajan et~al\mbox{.}(2024)]%
        {DBLP:journals/corr/abs-2405-01886}
\bibfield{author}{\bibinfo{person}{Ashwin~Kumar Gururajan}, \bibinfo{person}{Enrique Lopez{-}Cuena}, \bibinfo{person}{Jordi Bayarri{-}Planas}, \bibinfo{person}{Adri{\'{a}}n Tormos}, \bibinfo{person}{Daniel Hinjos}, \bibinfo{person}{Pablo Bernabeu{-}Perez}, \bibinfo{person}{Anna Arias{-}Duart}, \bibinfo{person}{Pablo~Agustin Martin{-}Torres}, \bibinfo{person}{Lucia Urcelay{-}Ganzabal}, \bibinfo{person}{Marta Gonzalez{-}Mallo}, \bibinfo{person}{Sergio {\'{A}}lvarez{-}Napagao}, \bibinfo{person}{Eduard~Ayguad{\'{e}} Parra}, {and} \bibinfo{person}{Ulises Cort{\'{e}}s~Dario Garcia{-}Gasulla}.} \bibinfo{year}{2024}\natexlab{}.
\newblock \showarticletitle{Aloe: {A} Family of Fine-tuned Open Healthcare LLMs}.
\newblock \bibinfo{journal}{\emph{CoRR}}  \bibinfo{volume}{abs/2405.01886} (\bibinfo{year}{2024}).
\newblock
\href{https://doi.org/10.48550/ARXIV.2405.01886}{doi:\nolinkurl{10.48550/ARXIV.2405.01886}}
\showeprint[arXiv]{2405.01886}


\bibitem[Han et~al\mbox{.}(2023)]%
        {DBLP:journals/corr/abs-2304-08247}
\bibfield{author}{\bibinfo{person}{Tianyu Han}, \bibinfo{person}{Lisa~C. Adams}, \bibinfo{person}{Jens{-}Michalis Papaioannou}, \bibinfo{person}{Paul Grundmann}, \bibinfo{person}{Tom Oberhauser}, \bibinfo{person}{Alexander L{\"{o}}ser}, \bibinfo{person}{Daniel Truhn}, {and} \bibinfo{person}{Keno~K. Bressem}.} \bibinfo{year}{2023}\natexlab{}.
\newblock \showarticletitle{MedAlpaca - An Open-Source Collection of Medical Conversational {AI} Models and Training Data}.
\newblock \bibinfo{journal}{\emph{CoRR}}  \bibinfo{volume}{abs/2304.08247} (\bibinfo{year}{2023}).
\newblock
\href{https://doi.org/10.48550/ARXIV.2304.08247}{doi:\nolinkurl{10.48550/ARXIV.2304.08247}}
\showeprint[arXiv]{2304.08247}


\bibitem[He et~al\mbox{.}(2020)]%
        {DBLP:conf/iclr/HeGSR20}
\bibfield{author}{\bibinfo{person}{Junxian He}, \bibinfo{person}{Jiatao Gu}, \bibinfo{person}{Jiajun Shen}, {and} \bibinfo{person}{Marc'Aurelio Ranzato}.} \bibinfo{year}{2020}\natexlab{}.
\newblock \showarticletitle{Revisiting Self-Training for Neural Sequence Generation}. In \bibinfo{booktitle}{\emph{8th International Conference on Learning Representations, {ICLR} 2020, Addis Ababa, Ethiopia, April 26-30, 2020}}. \bibinfo{publisher}{OpenReview.net}.
\newblock
\urldef\tempurl%
\url{https://openreview.net/forum?id=SJgdnAVKDH}
\showURL{%
\tempurl}


\bibitem[Held and Habash(2019)]%
        {DBLP:conf/acl/HeldH19}
\bibfield{author}{\bibinfo{person}{William Held} {and} \bibinfo{person}{Nizar Habash}.} \bibinfo{year}{2019}\natexlab{}.
\newblock \showarticletitle{The Effectiveness of Simple Hybrid Systems for Hypernym Discovery}. In \bibinfo{booktitle}{\emph{Proceedings of the 57th Conference of the Association for Computational Linguistics, {ACL} 2019, Florence, Italy, July 28- August 2, 2019, Volume 1: Long Papers}}, \bibfield{editor}{\bibinfo{person}{Anna Korhonen}, \bibinfo{person}{David~R. Traum}, {and} \bibinfo{person}{Llu{\'{\i}}s M{\`{a}}rquez}} (Eds.). \bibinfo{publisher}{Association for Computational Linguistics}, \bibinfo{pages}{3362--3367}.
\newblock
\href{https://doi.org/10.18653/V1/P19-1327}{doi:\nolinkurl{10.18653/V1/P19-1327}}


\bibitem[Hendrycks et~al\mbox{.}(2021)]%
        {DBLP:conf/iclr/HendrycksBBZMSS21}
\bibfield{author}{\bibinfo{person}{Dan Hendrycks}, \bibinfo{person}{Collin Burns}, \bibinfo{person}{Steven Basart}, \bibinfo{person}{Andy Zou}, \bibinfo{person}{Mantas Mazeika}, \bibinfo{person}{Dawn Song}, {and} \bibinfo{person}{Jacob Steinhardt}.} \bibinfo{year}{2021}\natexlab{}.
\newblock \showarticletitle{Measuring Massive Multitask Language Understanding}. In \bibinfo{booktitle}{\emph{9th International Conference on Learning Representations, {ICLR} 2021, Virtual Event, Austria, May 3-7, 2021}}. \bibinfo{publisher}{OpenReview.net}.
\newblock
\urldef\tempurl%
\url{https://openreview.net/forum?id=d7KBjmI3GmQ}
\showURL{%
\tempurl}


\bibitem[Hosseini et~al\mbox{.}(2024)]%
        {DBLP:journals/corr/abs-2402-06457}
\bibfield{author}{\bibinfo{person}{Arian Hosseini}, \bibinfo{person}{Xingdi Yuan}, \bibinfo{person}{Nikolay Malkin}, \bibinfo{person}{Aaron~C. Courville}, \bibinfo{person}{Alessandro Sordoni}, {and} \bibinfo{person}{Rishabh Agarwal}.} \bibinfo{year}{2024}\natexlab{}.
\newblock \showarticletitle{V-STaR: Training Verifiers for Self-Taught Reasoners}.
\newblock \bibinfo{journal}{\emph{CoRR}}  \bibinfo{volume}{abs/2402.06457} (\bibinfo{year}{2024}).
\newblock
\href{https://doi.org/10.48550/ARXIV.2402.06457}{doi:\nolinkurl{10.48550/ARXIV.2402.06457}}
\showeprint[arXiv]{2402.06457}


\bibitem[Hu et~al\mbox{.}(2022)]%
        {DBLP:conf/iclr/HuSWALWWC22}
\bibfield{author}{\bibinfo{person}{Edward~J. Hu}, \bibinfo{person}{Yelong Shen}, \bibinfo{person}{Phillip Wallis}, \bibinfo{person}{Zeyuan Allen{-}Zhu}, \bibinfo{person}{Yuanzhi Li}, \bibinfo{person}{Shean Wang}, \bibinfo{person}{Lu Wang}, {and} \bibinfo{person}{Weizhu Chen}.} \bibinfo{year}{2022}\natexlab{}.
\newblock \showarticletitle{LoRA: Low-Rank Adaptation of Large Language Models}. In \bibinfo{booktitle}{\emph{The Tenth International Conference on Learning Representations, {ICLR} 2022, Virtual Event, April 25-29, 2022}}. \bibinfo{publisher}{OpenReview.net}.
\newblock
\urldef\tempurl%
\url{https://openreview.net/forum?id=nZeVKeeFYf9}
\showURL{%
\tempurl}


\bibitem[Huang et~al\mbox{.}(2024)]%
        {DBLP:conf/iclr/0009CMZYSZ24}
\bibfield{author}{\bibinfo{person}{Jie Huang}, \bibinfo{person}{Xinyun Chen}, \bibinfo{person}{Swaroop Mishra}, \bibinfo{person}{Huaixiu~Steven Zheng}, \bibinfo{person}{Adams~Wei Yu}, \bibinfo{person}{Xinying Song}, {and} \bibinfo{person}{Denny Zhou}.} \bibinfo{year}{2024}\natexlab{}.
\newblock \showarticletitle{Large Language Models Cannot Self-Correct Reasoning Yet}. In \bibinfo{booktitle}{\emph{The Twelfth International Conference on Learning Representations, {ICLR} 2024, Vienna, Austria, May 7-11, 2024}}. \bibinfo{publisher}{OpenReview.net}.
\newblock
\urldef\tempurl%
\url{https://openreview.net/forum?id=IkmD3fKBPQ}
\showURL{%
\tempurl}


\bibitem[Huang et~al\mbox{.}(2023a)]%
        {DBLP:conf/emnlp/0001GHW00023}
\bibfield{author}{\bibinfo{person}{Jiaxin Huang}, \bibinfo{person}{Shixiang Gu}, \bibinfo{person}{Le Hou}, \bibinfo{person}{Yuexin Wu}, \bibinfo{person}{Xuezhi Wang}, \bibinfo{person}{Hongkun Yu}, {and} \bibinfo{person}{Jiawei Han}.} \bibinfo{year}{2023}\natexlab{a}.
\newblock \showarticletitle{Large Language Models Can Self-Improve}. In \bibinfo{booktitle}{\emph{Proceedings of the 2023 Conference on Empirical Methods in Natural Language Processing, {EMNLP} 2023, Singapore, December 6-10, 2023}}, \bibfield{editor}{\bibinfo{person}{Houda Bouamor}, \bibinfo{person}{Juan Pino}, {and} \bibinfo{person}{Kalika Bali}} (Eds.). \bibinfo{publisher}{Association for Computational Linguistics}, \bibinfo{pages}{1051--1068}.
\newblock
\href{https://doi.org/10.18653/V1/2023.EMNLP-MAIN.67}{doi:\nolinkurl{10.18653/V1/2023.EMNLP-MAIN.67}}


\bibitem[Huang et~al\mbox{.}(2023b)]%
        {DBLP:conf/acl/0002WHZS0WLFS023}
\bibfield{author}{\bibinfo{person}{Zijie Huang}, \bibinfo{person}{Daheng Wang}, \bibinfo{person}{Binxuan Huang}, \bibinfo{person}{Chenwei Zhang}, \bibinfo{person}{Jingbo Shang}, \bibinfo{person}{Yan Liang}, \bibinfo{person}{Zhengyang Wang}, \bibinfo{person}{Xian Li}, \bibinfo{person}{Christos Faloutsos}, \bibinfo{person}{Yizhou Sun}, {and} \bibinfo{person}{Wei Wang}.} \bibinfo{year}{2023}\natexlab{b}.
\newblock \showarticletitle{Concept2Box: Joint Geometric Embeddings for Learning Two-View Knowledge Graphs}. In \bibinfo{booktitle}{\emph{Findings of the Association for Computational Linguistics: {ACL} 2023, Toronto, Canada, July 9-14, 2023}}, \bibfield{editor}{\bibinfo{person}{Anna Rogers}, \bibinfo{person}{Jordan~L. Boyd{-}Graber}, {and} \bibinfo{person}{Naoaki Okazaki}} (Eds.). \bibinfo{publisher}{Association for Computational Linguistics}, \bibinfo{pages}{10105--10118}.
\newblock
\href{https://doi.org/10.18653/V1/2023.FINDINGS-ACL.642}{doi:\nolinkurl{10.18653/V1/2023.FINDINGS-ACL.642}}


\bibitem[Jiang et~al\mbox{.}(2023)]%
        {DBLP:journals/corr/abs-2310-06825}
\bibfield{author}{\bibinfo{person}{Albert~Q. Jiang}, \bibinfo{person}{Alexandre Sablayrolles}, \bibinfo{person}{Arthur Mensch}, \bibinfo{person}{Chris Bamford}, \bibinfo{person}{Devendra~Singh Chaplot}, \bibinfo{person}{Diego de Las~Casas}, \bibinfo{person}{Florian Bressand}, \bibinfo{person}{Gianna Lengyel}, \bibinfo{person}{Guillaume Lample}, \bibinfo{person}{Lucile Saulnier}, \bibinfo{person}{L{\'{e}}lio~Renard Lavaud}, \bibinfo{person}{Marie{-}Anne Lachaux}, \bibinfo{person}{Pierre Stock}, \bibinfo{person}{Teven~Le Scao}, \bibinfo{person}{Thibaut Lavril}, \bibinfo{person}{Thomas Wang}, \bibinfo{person}{Timoth{\'{e}}e Lacroix}, {and} \bibinfo{person}{William~El Sayed}.} \bibinfo{year}{2023}\natexlab{}.
\newblock \showarticletitle{Mistral 7B}.
\newblock \bibinfo{journal}{\emph{CoRR}}  \bibinfo{volume}{abs/2310.06825} (\bibinfo{year}{2023}).
\newblock
\href{https://doi.org/10.48550/ARXIV.2310.06825}{doi:\nolinkurl{10.48550/ARXIV.2310.06825}}
\showeprint[arXiv]{2310.06825}


\bibitem[Jin et~al\mbox{.}(2020)]%
        {DBLP:journals/corr/abs-2009-13081}
\bibfield{author}{\bibinfo{person}{Di Jin}, \bibinfo{person}{Eileen Pan}, \bibinfo{person}{Nassim Oufattole}, \bibinfo{person}{Wei{-}Hung Weng}, \bibinfo{person}{Hanyi Fang}, {and} \bibinfo{person}{Peter Szolovits}.} \bibinfo{year}{2020}\natexlab{}.
\newblock \showarticletitle{What Disease does this Patient Have? {A} Large-scale Open Domain Question Answering Dataset from Medical Exams}.
\newblock \bibinfo{journal}{\emph{CoRR}}  \bibinfo{volume}{abs/2009.13081} (\bibinfo{year}{2020}).
\newblock
\showeprint[arXiv]{2009.13081}
\urldef\tempurl%
\url{https://arxiv.org/abs/2009.13081}
\showURL{%
\tempurl}


\bibitem[Jin et~al\mbox{.}(2019)]%
        {DBLP:conf/emnlp/JinDLCL19}
\bibfield{author}{\bibinfo{person}{Qiao Jin}, \bibinfo{person}{Bhuwan Dhingra}, \bibinfo{person}{Zhengping Liu}, \bibinfo{person}{William~W. Cohen}, {and} \bibinfo{person}{Xinghua Lu}.} \bibinfo{year}{2019}\natexlab{}.
\newblock \showarticletitle{PubMedQA: {A} Dataset for Biomedical Research Question Answering}. In \bibinfo{booktitle}{\emph{Proceedings of the 2019 Conference on Empirical Methods in Natural Language Processing and the 9th International Joint Conference on Natural Language Processing, {EMNLP-IJCNLP} 2019, Hong Kong, China, November 3-7, 2019}}, \bibfield{editor}{\bibinfo{person}{Kentaro Inui}, \bibinfo{person}{Jing Jiang}, \bibinfo{person}{Vincent Ng}, {and} \bibinfo{person}{Xiaojun Wan}} (Eds.). \bibinfo{publisher}{Association for Computational Linguistics}, \bibinfo{pages}{2567--2577}.
\newblock
\href{https://doi.org/10.18653/V1/D19-1259}{doi:\nolinkurl{10.18653/V1/D19-1259}}


\bibitem[Joshi et~al\mbox{.}(2017)]%
        {DBLP:conf/acl/JoshiCWZ17}
\bibfield{author}{\bibinfo{person}{Mandar Joshi}, \bibinfo{person}{Eunsol Choi}, \bibinfo{person}{Daniel~S. Weld}, {and} \bibinfo{person}{Luke Zettlemoyer}.} \bibinfo{year}{2017}\natexlab{}.
\newblock \showarticletitle{TriviaQA: {A} Large Scale Distantly Supervised Challenge Dataset for Reading Comprehension}. In \bibinfo{booktitle}{\emph{Proceedings of the 55th Annual Meeting of the Association for Computational Linguistics, {ACL} 2017, Vancouver, Canada, July 30 - August 4, Volume 1: Long Papers}}, \bibfield{editor}{\bibinfo{person}{Regina Barzilay} {and} \bibinfo{person}{Min{-}Yen Kan}} (Eds.). \bibinfo{publisher}{Association for Computational Linguistics}, \bibinfo{pages}{1601--1611}.
\newblock
\href{https://doi.org/10.18653/V1/P17-1147}{doi:\nolinkurl{10.18653/V1/P17-1147}}


\bibitem[Labrak et~al\mbox{.}(2024)]%
        {DBLP:conf/acl/LabrakBMGRD24}
\bibfield{author}{\bibinfo{person}{Yanis Labrak}, \bibinfo{person}{Adrien Bazoge}, \bibinfo{person}{Emmanuel Morin}, \bibinfo{person}{Pierre{-}Antoine Gourraud}, \bibinfo{person}{Mickael Rouvier}, {and} \bibinfo{person}{Richard Dufour}.} \bibinfo{year}{2024}\natexlab{}.
\newblock \showarticletitle{BioMistral: {A} Collection of Open-Source Pretrained Large Language Models for Medical Domains}. In \bibinfo{booktitle}{\emph{Findings of the Association for Computational Linguistics, {ACL} 2024, Bangkok, Thailand and virtual meeting, August 11-16, 2024}}, \bibfield{editor}{\bibinfo{person}{Lun{-}Wei Ku}, \bibinfo{person}{Andre Martins}, {and} \bibinfo{person}{Vivek Srikumar}} (Eds.). \bibinfo{publisher}{Association for Computational Linguistics}, \bibinfo{pages}{5848--5864}.
\newblock
\urldef\tempurl%
\url{https://aclanthology.org/2024.findings-acl.348}
\showURL{%
\tempurl}


\bibitem[Li et~al\mbox{.}(2024)]%
        {DBLP:conf/acl/LiYBZLSLSYWLXBF24}
\bibfield{author}{\bibinfo{person}{Jiawei Li}, \bibinfo{person}{Yizhe Yang}, \bibinfo{person}{Yu Bai}, \bibinfo{person}{Xiaofeng Zhou}, \bibinfo{person}{Yinghao Li}, \bibinfo{person}{Huashan Sun}, \bibinfo{person}{Yuhang Liu}, \bibinfo{person}{Xingpeng Si}, \bibinfo{person}{Yuhao Ye}, \bibinfo{person}{Yixiao Wu}, \bibinfo{person}{Yiguan Lin}, \bibinfo{person}{Bin Xu}, \bibinfo{person}{Ren Bowen}, \bibinfo{person}{Chong Feng}, \bibinfo{person}{Yang Gao}, {and} \bibinfo{person}{Heyan Huang}.} \bibinfo{year}{2024}\natexlab{}.
\newblock \showarticletitle{Fundamental Capabilities of Large Language Models and their Applications in Domain Scenarios: {A} Survey}. In \bibinfo{booktitle}{\emph{Proceedings of the 62nd Annual Meeting of the Association for Computational Linguistics (Volume 1: Long Papers), {ACL} 2024, Bangkok, Thailand, August 11-16, 2024}}, \bibfield{editor}{\bibinfo{person}{Lun{-}Wei Ku}, \bibinfo{person}{Andre Martins}, {and} \bibinfo{person}{Vivek Srikumar}} (Eds.). \bibinfo{publisher}{Association for Computational Linguistics}, \bibinfo{pages}{11116--11141}.
\newblock
\urldef\tempurl%
\url{https://aclanthology.org/2024.acl-long.599}
\showURL{%
\tempurl}


\bibitem[Li et~al\mbox{.}(2023)]%
        {alpaca_eval}
\bibfield{author}{\bibinfo{person}{Xuechen Li}, \bibinfo{person}{Tianyi Zhang}, \bibinfo{person}{Yann Dubois}, \bibinfo{person}{Rohan Taori}, \bibinfo{person}{Ishaan Gulrajani}, \bibinfo{person}{Carlos Guestrin}, \bibinfo{person}{Percy Liang}, {and} \bibinfo{person}{Tatsunori~B. Hashimoto}.} \bibinfo{year}{2023}\natexlab{}.
\newblock \bibinfo{title}{AlpacaEval: An Automatic Evaluator of Instruction-following Models}.
\newblock \bibinfo{howpublished}{\url{https://github.com/tatsu-lab/alpaca_eval}}.
\newblock


\bibitem[Lin et~al\mbox{.}(2024)]%
        {DBLP:journals/corr/abs-2404-07965}
\bibfield{author}{\bibinfo{person}{Zhenghao Lin}, \bibinfo{person}{Zhibin Gou}, \bibinfo{person}{Yeyun Gong}, \bibinfo{person}{Xiao Liu}, \bibinfo{person}{Yelong Shen}, \bibinfo{person}{Ruochen Xu}, \bibinfo{person}{Chen Lin}, \bibinfo{person}{Yujiu Yang}, \bibinfo{person}{Jian Jiao}, \bibinfo{person}{Nan Duan}, {and} \bibinfo{person}{Weizhu Chen}.} \bibinfo{year}{2024}\natexlab{}.
\newblock \showarticletitle{Rho-1: Not All Tokens Are What You Need}.
\newblock \bibinfo{journal}{\emph{CoRR}}  \bibinfo{volume}{abs/2404.07965} (\bibinfo{year}{2024}).
\newblock
\href{https://doi.org/10.48550/ARXIV.2404.07965}{doi:\nolinkurl{10.48550/ARXIV.2404.07965}}
\showeprint[arXiv]{2404.07965}


\bibitem[Luo et~al\mbox{.}(2022)]%
        {DBLP:journals/bib/LuoSXQZPL22}
\bibfield{author}{\bibinfo{person}{Renqian Luo}, \bibinfo{person}{Liai Sun}, \bibinfo{person}{Yingce Xia}, \bibinfo{person}{Tao Qin}, \bibinfo{person}{Sheng Zhang}, \bibinfo{person}{Hoifung Poon}, {and} \bibinfo{person}{Tie{-}Yan Liu}.} \bibinfo{year}{2022}\natexlab{}.
\newblock \showarticletitle{BioGPT: generative pre-trained transformer for biomedical text generation and mining}.
\newblock \bibinfo{journal}{\emph{Briefings Bioinform.}} \bibinfo{volume}{23}, \bibinfo{number}{6} (\bibinfo{year}{2022}).
\newblock
\href{https://doi.org/10.1093/BIB/BBAC409}{doi:\nolinkurl{10.1093/BIB/BBAC409}}


\bibitem[Meng et~al\mbox{.}(2023)]%
        {DBLP:conf/icml/MengMHZA023}
\bibfield{author}{\bibinfo{person}{Yu Meng}, \bibinfo{person}{Martin Michalski}, \bibinfo{person}{Jiaxin Huang}, \bibinfo{person}{Yu Zhang}, \bibinfo{person}{Tarek~F. Abdelzaher}, {and} \bibinfo{person}{Jiawei Han}.} \bibinfo{year}{2023}\natexlab{}.
\newblock \showarticletitle{Tuning Language Models as Training Data Generators for Augmentation-Enhanced Few-Shot Learning}. In \bibinfo{booktitle}{\emph{International Conference on Machine Learning, {ICML} 2023, 23-29 July 2023, Honolulu, Hawaii, {USA}}} \emph{(\bibinfo{series}{Proceedings of Machine Learning Research}, Vol.~\bibinfo{volume}{202})}, \bibfield{editor}{\bibinfo{person}{Andreas Krause}, \bibinfo{person}{Emma Brunskill}, \bibinfo{person}{Kyunghyun Cho}, \bibinfo{person}{Barbara Engelhardt}, \bibinfo{person}{Sivan Sabato}, {and} \bibinfo{person}{Jonathan Scarlett}} (Eds.). \bibinfo{publisher}{{PMLR}}, \bibinfo{pages}{24457--24477}.
\newblock
\urldef\tempurl%
\url{https://proceedings.mlr.press/v202/meng23b.html}
\showURL{%
\tempurl}


\bibitem[Miller(1994)]%
        {DBLP:conf/naacl/Miller94a}
\bibfield{author}{\bibinfo{person}{George~A. Miller}.} \bibinfo{year}{1994}\natexlab{}.
\newblock \showarticletitle{{WORDNET:} {A} Lexical Database for English}. In \bibinfo{booktitle}{\emph{Human Language Technology, Proceedings of a Workshop held at Plainsboro, New Jerey, USA, March 8-11, 1994}}. \bibinfo{publisher}{Morgan Kaufmann}.
\newblock
\urldef\tempurl%
\url{https://aclanthology.org/H94-1111/}
\showURL{%
\tempurl}


\bibitem[Moskvoretskii et~al\mbox{.}(2024a)]%
        {DBLP:conf/acl/MoskvoretskiiNL24}
\bibfield{author}{\bibinfo{person}{Viktor Moskvoretskii}, \bibinfo{person}{Ekaterina Neminova}, \bibinfo{person}{Alina Lobanova}, \bibinfo{person}{Alexander Panchenko}, {and} \bibinfo{person}{Irina Nikishina}.} \bibinfo{year}{2024}\natexlab{a}.
\newblock \showarticletitle{TaxoLLaMA: WordNet-based Model for Solving Multiple Lexical Semantic Tasks}. In \bibinfo{booktitle}{\emph{Proceedings of the 62nd Annual Meeting of the Association for Computational Linguistics (Volume 1: Long Papers), {ACL} 2024, Bangkok, Thailand, August 11-16, 2024}}, \bibfield{editor}{\bibinfo{person}{Lun{-}Wei Ku}, \bibinfo{person}{Andre Martins}, {and} \bibinfo{person}{Vivek Srikumar}} (Eds.). \bibinfo{publisher}{Association for Computational Linguistics}, \bibinfo{pages}{2331--2350}.
\newblock
\urldef\tempurl%
\url{https://aclanthology.org/2024.acl-long.127}
\showURL{%
\tempurl}


\bibitem[Moskvoretskii et~al\mbox{.}(2024b)]%
        {DBLP:conf/coling/MoskvoretskiiPN24}
\bibfield{author}{\bibinfo{person}{Viktor Moskvoretskii}, \bibinfo{person}{Alexander Panchenko}, {and} \bibinfo{person}{Irina Nikishina}.} \bibinfo{year}{2024}\natexlab{b}.
\newblock \showarticletitle{Are Large Language Models Good at Lexical Semantics? {A} Case of Taxonomy Learning}. In \bibinfo{booktitle}{\emph{Proceedings of the 2024 Joint International Conference on Computational Linguistics, Language Resources and Evaluation, {LREC/COLING} 2024, 20-25 May, 2024, Torino, Italy}}, \bibfield{editor}{\bibinfo{person}{Nicoletta Calzolari}, \bibinfo{person}{Min{-}Yen Kan}, \bibinfo{person}{V{\'{e}}ronique Hoste}, \bibinfo{person}{Alessandro Lenci}, \bibinfo{person}{Sakriani Sakti}, {and} \bibinfo{person}{Nianwen Xue}} (Eds.). \bibinfo{publisher}{{ELRA} and {ICCL}}, \bibinfo{pages}{1498--1510}.
\newblock
\urldef\tempurl%
\url{https://aclanthology.org/2024.lrec-main.133}
\showURL{%
\tempurl}


\bibitem[Nikishina et~al\mbox{.}(2023)]%
        {DBLP:conf/ijcnlp/NikishinaCDPB23}
\bibfield{author}{\bibinfo{person}{Irina Nikishina}, \bibinfo{person}{Polina Chernomorchenko}, \bibinfo{person}{Anastasiia Demidova}, \bibinfo{person}{Alexander Panchenko}, {and} \bibinfo{person}{Chris Biemann}.} \bibinfo{year}{2023}\natexlab{}.
\newblock \showarticletitle{Predicting Terms in {IS-A} Relations with Pre-trained Transformers}. In \bibinfo{booktitle}{\emph{Findings of the Association for Computational Linguistics: {IJCNLP-AACL} 2023 - Findings, Nusa Dua, Bali, November 1-4, 2023}}, \bibfield{editor}{\bibinfo{person}{Jong~C. Park}, \bibinfo{person}{Yuki Arase}, \bibinfo{person}{Baotian Hu}, \bibinfo{person}{Wei Lu}, \bibinfo{person}{Derry Wijaya}, \bibinfo{person}{Ayu Purwarianti}, {and} \bibinfo{person}{Adila~Alfa Krisnadhi}} (Eds.). \bibinfo{publisher}{Association for Computational Linguistics}, \bibinfo{pages}{134--148}.
\newblock
\href{https://doi.org/10.18653/V1/2023.FINDINGS-IJCNLP.12}{doi:\nolinkurl{10.18653/V1/2023.FINDINGS-IJCNLP.12}}


\bibitem[OpenAI(2023)]%
        {DBLP:journals/corr/abs-2303-08774}
\bibfield{author}{\bibinfo{person}{OpenAI}.} \bibinfo{year}{2023}\natexlab{}.
\newblock \showarticletitle{{GPT-4} Technical Report}.
\newblock \bibinfo{journal}{\emph{CoRR}}  \bibinfo{volume}{abs/2303.08774} (\bibinfo{year}{2023}).
\newblock
\href{https://doi.org/10.48550/ARXIV.2303.08774}{doi:\nolinkurl{10.48550/ARXIV.2303.08774}}
\showeprint[arXiv]{2303.08774}


\bibitem[Pal et~al\mbox{.}(2022)]%
        {DBLP:conf/chil/PalUS22}
\bibfield{author}{\bibinfo{person}{Ankit Pal}, \bibinfo{person}{Logesh~Kumar Umapathi}, {and} \bibinfo{person}{Malaikannan Sankarasubbu}.} \bibinfo{year}{2022}\natexlab{}.
\newblock \showarticletitle{MedMCQA: {A} Large-scale Multi-Subject Multi-Choice Dataset for Medical domain Question Answering}. In \bibinfo{booktitle}{\emph{Conference on Health, Inference, and Learning, {CHIL} 2022, 7-8 April 2022, Virtual Event}} \emph{(\bibinfo{series}{Proceedings of Machine Learning Research}, Vol.~\bibinfo{volume}{174})}, \bibfield{editor}{\bibinfo{person}{Gerardo Flores}, \bibinfo{person}{George~H. Chen}, \bibinfo{person}{Tom~J. Pollard}, \bibinfo{person}{Joyce~C. Ho}, {and} \bibinfo{person}{Tristan Naumann}} (Eds.). \bibinfo{publisher}{{PMLR}}, \bibinfo{pages}{248--260}.
\newblock
\urldef\tempurl%
\url{https://proceedings.mlr.press/v174/pal22a.html}
\showURL{%
\tempurl}


\bibitem[Qi et~al\mbox{.}(2024)]%
        {DBLP:conf/iclr/Qi0XC0M024}
\bibfield{author}{\bibinfo{person}{Xiangyu Qi}, \bibinfo{person}{Yi Zeng}, \bibinfo{person}{Tinghao Xie}, \bibinfo{person}{Pin{-}Yu Chen}, \bibinfo{person}{Ruoxi Jia}, \bibinfo{person}{Prateek Mittal}, {and} \bibinfo{person}{Peter Henderson}.} \bibinfo{year}{2024}\natexlab{}.
\newblock \showarticletitle{Fine-tuning Aligned Language Models Compromises Safety, Even When Users Do Not Intend To!}. In \bibinfo{booktitle}{\emph{The Twelfth International Conference on Learning Representations, {ICLR} 2024, Vienna, Austria, May 7-11, 2024}}. \bibinfo{publisher}{OpenReview.net}.
\newblock
\urldef\tempurl%
\url{https://openreview.net/forum?id=hTEGyKf0dZ}
\showURL{%
\tempurl}


\bibitem[Raffel et~al\mbox{.}(2020)]%
        {DBLP:journals/jmlr/RaffelSRLNMZLL20}
\bibfield{author}{\bibinfo{person}{Colin Raffel}, \bibinfo{person}{Noam Shazeer}, \bibinfo{person}{Adam Roberts}, \bibinfo{person}{Katherine Lee}, \bibinfo{person}{Sharan Narang}, \bibinfo{person}{Michael Matena}, \bibinfo{person}{Yanqi Zhou}, \bibinfo{person}{Wei Li}, {and} \bibinfo{person}{Peter~J. Liu}.} \bibinfo{year}{2020}\natexlab{}.
\newblock \showarticletitle{Exploring the Limits of Transfer Learning with a Unified Text-to-Text Transformer}.
\newblock \bibinfo{journal}{\emph{J. Mach. Learn. Res.}}  \bibinfo{volume}{21} (\bibinfo{year}{2020}), \bibinfo{pages}{140:1--140:67}.
\newblock
\urldef\tempurl%
\url{http://jmlr.org/papers/v21/20-074.html}
\showURL{%
\tempurl}


\bibitem[Ren et~al\mbox{.}(2024)]%
        {DBLP:conf/acl/RenCLLHZWC024}
\bibfield{author}{\bibinfo{person}{Mengjie Ren}, \bibinfo{person}{Boxi Cao}, \bibinfo{person}{Hongyu Lin}, \bibinfo{person}{Cao Liu}, \bibinfo{person}{Xianpei Han}, \bibinfo{person}{Ke Zeng}, \bibinfo{person}{Guanglu Wan}, \bibinfo{person}{Xunliang Cai}, {and} \bibinfo{person}{Le Sun}.} \bibinfo{year}{2024}\natexlab{}.
\newblock \showarticletitle{Learning or Self-aligning? Rethinking Instruction Fine-tuning}. In \bibinfo{booktitle}{\emph{Proceedings of the 62nd Annual Meeting of the Association for Computational Linguistics (Volume 1: Long Papers), {ACL} 2024, Bangkok, Thailand, August 11-16, 2024}}, \bibfield{editor}{\bibinfo{person}{Lun{-}Wei Ku}, \bibinfo{person}{Andre Martins}, {and} \bibinfo{person}{Vivek Srikumar}} (Eds.). \bibinfo{publisher}{Association for Computational Linguistics}, \bibinfo{pages}{6090--6105}.
\newblock
\href{https://doi.org/10.18653/V1/2024.ACL-LONG.330}{doi:\nolinkurl{10.18653/V1/2024.ACL-LONG.330}}


\bibitem[Schulz and Klein(2008)]%
        {DBLP:journals/midm/SchulzK08}
\bibfield{author}{\bibinfo{person}{Stefan Schulz} {and} \bibinfo{person}{Gunnar~O. Klein}.} \bibinfo{year}{2008}\natexlab{}.
\newblock \showarticletitle{{SNOMED} {CT} - advances in concept mapping, retrieval, and ontological foundations. Selected contributions to the Semantic Mining Conference on {SNOMED} {CT} {(SMCS} 2006)}.
\newblock \bibinfo{journal}{\emph{{BMC} Medical Informatics Decis. Mak.}} \bibinfo{volume}{8}, \bibinfo{number}{{S-1}} (\bibinfo{year}{2008}), \bibinfo{pages}{S1}.
\newblock
\href{https://doi.org/10.1186/1472-6947-8-S1-S1}{doi:\nolinkurl{10.1186/1472-6947-8-S1-S1}}


\bibitem[Singh et~al\mbox{.}(2024)]%
        {DBLP:journals/tmlr/SinghCAAPGLH0XP24}
\bibfield{author}{\bibinfo{person}{Avi Singh}, \bibinfo{person}{John~D. Co{-}Reyes}, \bibinfo{person}{Rishabh Agarwal}, \bibinfo{person}{Ankesh Anand}, \bibinfo{person}{Piyush Patil}, \bibinfo{person}{Xavier Garcia}, \bibinfo{person}{Peter~J. Liu}, \bibinfo{person}{James Harrison}, \bibinfo{person}{Jaehoon Lee}, \bibinfo{person}{Kelvin Xu}, \bibinfo{person}{Aaron~T. Parisi}, \bibinfo{person}{Abhishek Kumar}, \bibinfo{person}{Alexander~A. Alemi}, \bibinfo{person}{Alex Rizkowsky}, \bibinfo{person}{Azade Nova}, \bibinfo{person}{Ben Adlam}, \bibinfo{person}{Bernd Bohnet}, \bibinfo{person}{Gamaleldin~Fathy Elsayed}, \bibinfo{person}{Hanie Sedghi}, \bibinfo{person}{Igor Mordatch}, \bibinfo{person}{Isabelle Simpson}, \bibinfo{person}{Izzeddin Gur}, \bibinfo{person}{Jasper Snoek}, \bibinfo{person}{Jeffrey Pennington}, \bibinfo{person}{Jiri Hron}, \bibinfo{person}{Kathleen Kenealy}, \bibinfo{person}{Kevin Swersky}, \bibinfo{person}{Kshiteej Mahajan}, \bibinfo{person}{Laura Culp}, \bibinfo{person}{Lechao Xiao},
  \bibinfo{person}{Maxwell~L. Bileschi}, \bibinfo{person}{Noah Constant}, \bibinfo{person}{Roman Novak}, \bibinfo{person}{Rosanne Liu}, \bibinfo{person}{Tris Warkentin}, \bibinfo{person}{Yundi Qian}, \bibinfo{person}{Yamini Bansal}, \bibinfo{person}{Ethan Dyer}, \bibinfo{person}{Behnam Neyshabur}, \bibinfo{person}{Jascha Sohl{-}Dickstein}, {and} \bibinfo{person}{Noah Fiedel}.} \bibinfo{year}{2024}\natexlab{}.
\newblock \showarticletitle{Beyond Human Data: Scaling Self-Training for Problem-Solving with Language Models}.
\newblock \bibinfo{journal}{\emph{Trans. Mach. Learn. Res.}}  \bibinfo{volume}{2024} (\bibinfo{year}{2024}).
\newblock
\urldef\tempurl%
\url{https://openreview.net/forum?id=lNAyUngGFK}
\showURL{%
\tempurl}


\bibitem[Touvron et~al\mbox{.}(2023)]%
        {DBLP:journals/corr/abs-2307-09288}
\bibfield{author}{\bibinfo{person}{Hugo Touvron}, \bibinfo{person}{Louis Martin}, \bibinfo{person}{Kevin Stone}, \bibinfo{person}{Peter Albert}, \bibinfo{person}{Amjad Almahairi}, \bibinfo{person}{Yasmine Babaei}, \bibinfo{person}{Nikolay Bashlykov}, \bibinfo{person}{Soumya Batra}, \bibinfo{person}{Prajjwal Bhargava}, \bibinfo{person}{Shruti Bhosale}, \bibinfo{person}{Dan Bikel}, \bibinfo{person}{Lukas Blecher}, \bibinfo{person}{Cristian Canton{-}Ferrer}, \bibinfo{person}{Moya Chen}, \bibinfo{person}{Guillem Cucurull}, \bibinfo{person}{David Esiobu}, \bibinfo{person}{Jude Fernandes}, \bibinfo{person}{Jeremy Fu}, \bibinfo{person}{Wenyin Fu}, \bibinfo{person}{Brian Fuller}, \bibinfo{person}{Cynthia Gao}, \bibinfo{person}{Vedanuj Goswami}, \bibinfo{person}{Naman Goyal}, \bibinfo{person}{Anthony Hartshorn}, \bibinfo{person}{Saghar Hosseini}, \bibinfo{person}{Rui Hou}, \bibinfo{person}{Hakan Inan}, \bibinfo{person}{Marcin Kardas}, \bibinfo{person}{Viktor Kerkez}, \bibinfo{person}{Madian Khabsa},
  \bibinfo{person}{Isabel Kloumann}, \bibinfo{person}{Artem Korenev}, \bibinfo{person}{Punit~Singh Koura}, \bibinfo{person}{Marie{-}Anne Lachaux}, \bibinfo{person}{Thibaut Lavril}, \bibinfo{person}{Jenya Lee}, \bibinfo{person}{Diana Liskovich}, \bibinfo{person}{Yinghai Lu}, \bibinfo{person}{Yuning Mao}, \bibinfo{person}{Xavier Martinet}, \bibinfo{person}{Todor Mihaylov}, \bibinfo{person}{Pushkar Mishra}, \bibinfo{person}{Igor Molybog}, \bibinfo{person}{Yixin Nie}, \bibinfo{person}{Andrew Poulton}, \bibinfo{person}{Jeremy Reizenstein}, \bibinfo{person}{Rashi Rungta}, \bibinfo{person}{Kalyan Saladi}, \bibinfo{person}{Alan Schelten}, \bibinfo{person}{Ruan Silva}, \bibinfo{person}{Eric~Michael Smith}, \bibinfo{person}{Ranjan Subramanian}, \bibinfo{person}{Xiaoqing~Ellen Tan}, \bibinfo{person}{Binh Tang}, \bibinfo{person}{Ross Taylor}, \bibinfo{person}{Adina Williams}, \bibinfo{person}{Jian~Xiang Kuan}, \bibinfo{person}{Puxin Xu}, \bibinfo{person}{Zheng Yan}, \bibinfo{person}{Iliyan Zarov}, \bibinfo{person}{Yuchen
  Zhang}, \bibinfo{person}{Angela Fan}, \bibinfo{person}{Melanie Kambadur}, \bibinfo{person}{Sharan Narang}, \bibinfo{person}{Aur{\'{e}}lien Rodriguez}, \bibinfo{person}{Robert Stojnic}, \bibinfo{person}{Sergey Edunov}, {and} \bibinfo{person}{Thomas Scialom}.} \bibinfo{year}{2023}\natexlab{}.
\newblock \showarticletitle{Llama 2: Open Foundation and Fine-Tuned Chat Models}.
\newblock \bibinfo{journal}{\emph{CoRR}}  \bibinfo{volume}{abs/2307.09288} (\bibinfo{year}{2023}).
\newblock
\href{https://doi.org/10.48550/ARXIV.2307.09288}{doi:\nolinkurl{10.48550/ARXIV.2307.09288}}
\showeprint[arXiv]{2307.09288}


\bibitem[Tyen et~al\mbox{.}(2024)]%
        {DBLP:conf/acl/TyenMCCM24}
\bibfield{author}{\bibinfo{person}{Gladys Tyen}, \bibinfo{person}{Hassan Mansoor}, \bibinfo{person}{Victor Carbune}, \bibinfo{person}{Peter Chen}, {and} \bibinfo{person}{Tony Mak}.} \bibinfo{year}{2024}\natexlab{}.
\newblock \showarticletitle{LLMs cannot find reasoning errors, but can correct them given the error location}. In \bibinfo{booktitle}{\emph{Findings of the Association for Computational Linguistics, {ACL} 2024, Bangkok, Thailand and virtual meeting, August 11-16, 2024}}, \bibfield{editor}{\bibinfo{person}{Lun{-}Wei Ku}, \bibinfo{person}{Andre Martins}, {and} \bibinfo{person}{Vivek Srikumar}} (Eds.). \bibinfo{publisher}{Association for Computational Linguistics}, \bibinfo{pages}{13894--13908}.
\newblock
\urldef\tempurl%
\url{https://aclanthology.org/2024.findings-acl.826}
\showURL{%
\tempurl}


\bibitem[Wang et~al\mbox{.}(2024a)]%
        {DBLP:journals/corr/abs-2406-14282}
\bibfield{author}{\bibinfo{person}{Junjie Wang}, \bibinfo{person}{Mingyang Chen}, \bibinfo{person}{Binbin Hu}, \bibinfo{person}{Dan Yang}, \bibinfo{person}{Ziqi Liu}, \bibinfo{person}{Yue Shen}, \bibinfo{person}{Peng Wei}, \bibinfo{person}{Zhiqiang Zhang}, \bibinfo{person}{Jinjie Gu}, \bibinfo{person}{Jun Zhou}, \bibinfo{person}{Jeff~Z. Pan}, \bibinfo{person}{Wen Zhang}, {and} \bibinfo{person}{Huajun Chen}.} \bibinfo{year}{2024}\natexlab{a}.
\newblock \showarticletitle{Learning to Plan for Retrieval-Augmented Large Language Models from Knowledge Graphs}.
\newblock \bibinfo{journal}{\emph{CoRR}}  \bibinfo{volume}{abs/2406.14282} (\bibinfo{year}{2024}).
\newblock
\href{https://doi.org/10.48550/ARXIV.2406.14282}{doi:\nolinkurl{10.48550/ARXIV.2406.14282}}
\showeprint[arXiv]{2406.14282}


\bibitem[Wang et~al\mbox{.}(2024b)]%
        {DBLP:journals/corr/abs-2407-15017}
\bibfield{author}{\bibinfo{person}{Mengru Wang}, \bibinfo{person}{Yunzhi Yao}, \bibinfo{person}{Ziwen Xu}, \bibinfo{person}{Shuofei Qiao}, \bibinfo{person}{Shumin Deng}, \bibinfo{person}{Peng Wang}, \bibinfo{person}{Xiang Chen}, \bibinfo{person}{Jia{-}Chen Gu}, \bibinfo{person}{Yong Jiang}, \bibinfo{person}{Pengjun Xie}, \bibinfo{person}{Fei Huang}, \bibinfo{person}{Huajun Chen}, {and} \bibinfo{person}{Ningyu Zhang}.} \bibinfo{year}{2024}\natexlab{b}.
\newblock \showarticletitle{Knowledge Mechanisms in Large Language Models: {A} Survey and Perspective}.
\newblock \bibinfo{journal}{\emph{CoRR}}  \bibinfo{volume}{abs/2407.15017} (\bibinfo{year}{2024}).
\newblock
\href{https://doi.org/10.48550/ARXIV.2407.15017}{doi:\nolinkurl{10.48550/ARXIV.2407.15017}}
\showeprint[arXiv]{2407.15017}


\bibitem[Wang et~al\mbox{.}(2021)]%
        {DBLP:conf/acl/WangBHDW21}
\bibfield{author}{\bibinfo{person}{Wenhui Wang}, \bibinfo{person}{Hangbo Bao}, \bibinfo{person}{Shaohan Huang}, \bibinfo{person}{Li Dong}, {and} \bibinfo{person}{Furu Wei}.} \bibinfo{year}{2021}\natexlab{}.
\newblock \showarticletitle{MiniLMv2: Multi-Head Self-Attention Relation Distillation for Compressing Pretrained Transformers}. In \bibinfo{booktitle}{\emph{Findings of the Association for Computational Linguistics: {ACL/IJCNLP} 2021, Online Event, August 1-6, 2021}} \emph{(\bibinfo{series}{Findings of {ACL}}, Vol.~\bibinfo{volume}{{ACL/IJCNLP} 2021})}, \bibfield{editor}{\bibinfo{person}{Chengqing Zong}, \bibinfo{person}{Fei Xia}, \bibinfo{person}{Wenjie Li}, {and} \bibinfo{person}{Roberto Navigli}} (Eds.). \bibinfo{publisher}{Association for Computational Linguistics}, \bibinfo{pages}{2140--2151}.
\newblock
\href{https://doi.org/10.18653/V1/2021.FINDINGS-ACL.188}{doi:\nolinkurl{10.18653/V1/2021.FINDINGS-ACL.188}}


\bibitem[Wang et~al\mbox{.}(2023)]%
        {DBLP:conf/acl/WangKMLSKH23}
\bibfield{author}{\bibinfo{person}{Yizhong Wang}, \bibinfo{person}{Yeganeh Kordi}, \bibinfo{person}{Swaroop Mishra}, \bibinfo{person}{Alisa Liu}, \bibinfo{person}{Noah~A. Smith}, \bibinfo{person}{Daniel Khashabi}, {and} \bibinfo{person}{Hannaneh Hajishirzi}.} \bibinfo{year}{2023}\natexlab{}.
\newblock \showarticletitle{Self-Instruct: Aligning Language Models with Self-Generated Instructions}. In \bibinfo{booktitle}{\emph{Proceedings of the 61st Annual Meeting of the Association for Computational Linguistics (Volume 1: Long Papers), {ACL} 2023, Toronto, Canada, July 9-14, 2023}}, \bibfield{editor}{\bibinfo{person}{Anna Rogers}, \bibinfo{person}{Jordan~L. Boyd{-}Graber}, {and} \bibinfo{person}{Naoaki Okazaki}} (Eds.). \bibinfo{publisher}{Association for Computational Linguistics}, \bibinfo{pages}{13484--13508}.
\newblock
\href{https://doi.org/10.18653/V1/2023.ACL-LONG.754}{doi:\nolinkurl{10.18653/V1/2023.ACL-LONG.754}}


\bibitem[Wu et~al\mbox{.}(2023b)]%
        {DBLP:journals/corr/abs-2304-14454}
\bibfield{author}{\bibinfo{person}{Chaoyi Wu}, \bibinfo{person}{Xiaoman Zhang}, \bibinfo{person}{Ya Zhang}, \bibinfo{person}{Yanfeng Wang}, {and} \bibinfo{person}{Weidi Xie}.} \bibinfo{year}{2023}\natexlab{b}.
\newblock \showarticletitle{PMC-LLaMA: Further Finetuning LLaMA on Medical Papers}.
\newblock \bibinfo{journal}{\emph{CoRR}}  \bibinfo{volume}{abs/2304.14454} (\bibinfo{year}{2023}).
\newblock
\href{https://doi.org/10.48550/ARXIV.2304.14454}{doi:\nolinkurl{10.48550/ARXIV.2304.14454}}
\showeprint[arXiv]{2304.14454}


\bibitem[Wu et~al\mbox{.}(2023a)]%
        {DBLP:journals/corr/abs-2303-17564}
\bibfield{author}{\bibinfo{person}{Shijie Wu}, \bibinfo{person}{Ozan Irsoy}, \bibinfo{person}{Steven Lu}, \bibinfo{person}{Vadim Dabravolski}, \bibinfo{person}{Mark Dredze}, \bibinfo{person}{Sebastian Gehrmann}, \bibinfo{person}{Prabhanjan Kambadur}, \bibinfo{person}{David~S. Rosenberg}, {and} \bibinfo{person}{Gideon Mann}.} \bibinfo{year}{2023}\natexlab{a}.
\newblock \showarticletitle{BloombergGPT: {A} Large Language Model for Finance}.
\newblock \bibinfo{journal}{\emph{CoRR}}  \bibinfo{volume}{abs/2303.17564} (\bibinfo{year}{2023}).
\newblock
\href{https://doi.org/10.48550/ARXIV.2303.17564}{doi:\nolinkurl{10.48550/ARXIV.2303.17564}}
\showeprint[arXiv]{2303.17564}


\bibitem[Xiao et~al\mbox{.}(2018)]%
        {DBLP:conf/ijcai/XiaoCKLPRZ18}
\bibfield{author}{\bibinfo{person}{Guohui Xiao}, \bibinfo{person}{Diego Calvanese}, \bibinfo{person}{Roman Kontchakov}, \bibinfo{person}{Domenico Lembo}, \bibinfo{person}{Antonella Poggi}, \bibinfo{person}{Riccardo Rosati}, {and} \bibinfo{person}{Michael Zakharyaschev}.} \bibinfo{year}{2018}\natexlab{}.
\newblock \showarticletitle{Ontology-Based Data Access: {A} Survey}. In \bibinfo{booktitle}{\emph{Proceedings of the Twenty-Seventh International Joint Conference on Artificial Intelligence, {IJCAI} 2018, July 13-19, 2018, Stockholm, Sweden}}, \bibfield{editor}{\bibinfo{person}{J{\'{e}}r{\^{o}}me Lang}} (Ed.). \bibinfo{publisher}{ijcai.org}, \bibinfo{pages}{5511--5519}.
\newblock
\href{https://doi.org/10.24963/IJCAI.2018/777}{doi:\nolinkurl{10.24963/IJCAI.2018/777}}


\bibitem[Xie et~al\mbox{.}(2020)]%
        {DBLP:conf/cvpr/XieLHL20}
\bibfield{author}{\bibinfo{person}{Qizhe Xie}, \bibinfo{person}{Minh{-}Thang Luong}, \bibinfo{person}{Eduard~H. Hovy}, {and} \bibinfo{person}{Quoc~V. Le}.} \bibinfo{year}{2020}\natexlab{}.
\newblock \showarticletitle{Self-Training With Noisy Student Improves ImageNet Classification}. In \bibinfo{booktitle}{\emph{2020 {IEEE/CVF} Conference on Computer Vision and Pattern Recognition, {CVPR} 2020, Seattle, WA, USA, June 13-19, 2020}}. \bibinfo{publisher}{Computer Vision Foundation / {IEEE}}, \bibinfo{pages}{10684--10695}.
\newblock
\href{https://doi.org/10.1109/CVPR42600.2020.01070}{doi:\nolinkurl{10.1109/CVPR42600.2020.01070}}


\bibitem[Yang et~al\mbox{.}(2024b)]%
        {DBLP:journals/corr/abs-2407-10671}
\bibfield{author}{\bibinfo{person}{An Yang}, \bibinfo{person}{Baosong Yang}, \bibinfo{person}{Binyuan Hui}, \bibinfo{person}{Bo Zheng}, \bibinfo{person}{Bowen Yu}, \bibinfo{person}{Chang Zhou}, \bibinfo{person}{Chengpeng Li}, \bibinfo{person}{Chengyuan Li}, \bibinfo{person}{Dayiheng Liu}, \bibinfo{person}{Fei Huang}, \bibinfo{person}{Guanting Dong}, \bibinfo{person}{Haoran Wei}, \bibinfo{person}{Huan Lin}, \bibinfo{person}{Jialong Tang}, \bibinfo{person}{Jialin Wang}, \bibinfo{person}{Jian Yang}, \bibinfo{person}{Jianhong Tu}, \bibinfo{person}{Jianwei Zhang}, \bibinfo{person}{Jianxin Ma}, \bibinfo{person}{Jianxin Yang}, \bibinfo{person}{Jin Xu}, \bibinfo{person}{Jingren Zhou}, \bibinfo{person}{Jinze Bai}, \bibinfo{person}{Jinzheng He}, \bibinfo{person}{Junyang Lin}, \bibinfo{person}{Kai Dang}, \bibinfo{person}{Keming Lu}, \bibinfo{person}{Keqin Chen}, \bibinfo{person}{Kexin Yang}, \bibinfo{person}{Mei Li}, \bibinfo{person}{Mingfeng Xue}, \bibinfo{person}{Na Ni}, \bibinfo{person}{Pei Zhang},
  \bibinfo{person}{Peng Wang}, \bibinfo{person}{Ru Peng}, \bibinfo{person}{Rui Men}, \bibinfo{person}{Ruize Gao}, \bibinfo{person}{Runji Lin}, \bibinfo{person}{Shijie Wang}, \bibinfo{person}{Shuai Bai}, \bibinfo{person}{Sinan Tan}, \bibinfo{person}{Tianhang Zhu}, \bibinfo{person}{Tianhao Li}, \bibinfo{person}{Tianyu Liu}, \bibinfo{person}{Wenbin Ge}, \bibinfo{person}{Xiaodong Deng}, \bibinfo{person}{Xiaohuan Zhou}, \bibinfo{person}{Xingzhang Ren}, \bibinfo{person}{Xinyu Zhang}, \bibinfo{person}{Xipin Wei}, \bibinfo{person}{Xuancheng Ren}, \bibinfo{person}{Xuejing Liu}, \bibinfo{person}{Yang Fan}, \bibinfo{person}{Yang Yao}, \bibinfo{person}{Yichang Zhang}, \bibinfo{person}{Yu Wan}, \bibinfo{person}{Yunfei Chu}, \bibinfo{person}{Yuqiong Liu}, \bibinfo{person}{Zeyu Cui}, \bibinfo{person}{Zhenru Zhang}, \bibinfo{person}{Zhifang Guo}, {and} \bibinfo{person}{Zhihao Fan}.} \bibinfo{year}{2024}\natexlab{b}.
\newblock \showarticletitle{Qwen2 Technical Report}.
\newblock \bibinfo{journal}{\emph{CoRR}}  \bibinfo{volume}{abs/2407.10671} (\bibinfo{year}{2024}).
\newblock
\href{https://doi.org/10.48550/ARXIV.2407.10671}{doi:\nolinkurl{10.48550/ARXIV.2407.10671}}
\showeprint[arXiv]{2407.10671}


\bibitem[Yang et~al\mbox{.}(2022)]%
        {DBLP:journals/corr/abs-2203-03540}
\bibfield{author}{\bibinfo{person}{Xi Yang}, \bibinfo{person}{Nima~M. Pournejatian}, \bibinfo{person}{Hoo~Chang Shin}, \bibinfo{person}{Kaleb~E. Smith}, \bibinfo{person}{Christopher Parisien}, \bibinfo{person}{Colin Compas}, \bibinfo{person}{Cheryl Martin}, \bibinfo{person}{Mona~G. Flores}, \bibinfo{person}{Ying Zhang}, \bibinfo{person}{Tanja Magoc}, \bibinfo{person}{Christopher~A. Harle}, \bibinfo{person}{Gloria~P. Lipori}, \bibinfo{person}{Duane~A. Mitchell}, \bibinfo{person}{William~R. Hogan}, \bibinfo{person}{Elizabeth~A. Shenkman}, \bibinfo{person}{Jiang Bian}, {and} \bibinfo{person}{Yonghui Wu}.} \bibinfo{year}{2022}\natexlab{}.
\newblock \showarticletitle{GatorTron: {A} Large Clinical Language Model to Unlock Patient Information from Unstructured Electronic Health Records}.
\newblock \bibinfo{journal}{\emph{CoRR}}  \bibinfo{volume}{abs/2203.03540} (\bibinfo{year}{2022}).
\newblock
\href{https://doi.org/10.48550/ARXIV.2203.03540}{doi:\nolinkurl{10.48550/ARXIV.2203.03540}}
\showeprint[arXiv]{2203.03540}


\bibitem[Yang et~al\mbox{.}(2024a)]%
        {DBLP:conf/acl/YangPFWCZL24}
\bibfield{author}{\bibinfo{person}{Zhaorui Yang}, \bibinfo{person}{Tianyu Pang}, \bibinfo{person}{Haozhe Feng}, \bibinfo{person}{Han Wang}, \bibinfo{person}{Wei Chen}, \bibinfo{person}{Minfeng Zhu}, {and} \bibinfo{person}{Qian Liu}.} \bibinfo{year}{2024}\natexlab{a}.
\newblock \showarticletitle{Self-Distillation Bridges Distribution Gap in Language Model Fine-Tuning}. In \bibinfo{booktitle}{\emph{Proceedings of the 62nd Annual Meeting of the Association for Computational Linguistics (Volume 1: Long Papers), {ACL} 2024, Bangkok, Thailand, August 11-16, 2024}}, \bibfield{editor}{\bibinfo{person}{Lun{-}Wei Ku}, \bibinfo{person}{Andre Martins}, {and} \bibinfo{person}{Vivek Srikumar}} (Eds.). \bibinfo{publisher}{Association for Computational Linguistics}, \bibinfo{pages}{1028--1043}.
\newblock
\urldef\tempurl%
\url{https://aclanthology.org/2024.acl-long.58}
\showURL{%
\tempurl}


\bibitem[Zelikman et~al\mbox{.}(2022)]%
        {DBLP:conf/nips/ZelikmanWMG22}
\bibfield{author}{\bibinfo{person}{Eric Zelikman}, \bibinfo{person}{Yuhuai Wu}, \bibinfo{person}{Jesse Mu}, {and} \bibinfo{person}{Noah~D. Goodman}.} \bibinfo{year}{2022}\natexlab{}.
\newblock \showarticletitle{STaR: Bootstrapping Reasoning With Reasoning}. In \bibinfo{booktitle}{\emph{Advances in Neural Information Processing Systems 35: Annual Conference on Neural Information Processing Systems 2022, NeurIPS 2022, New Orleans, LA, USA, November 28 - December 9, 2022}}, \bibfield{editor}{\bibinfo{person}{Sanmi Koyejo}, \bibinfo{person}{S.~Mohamed}, \bibinfo{person}{A.~Agarwal}, \bibinfo{person}{Danielle Belgrave}, \bibinfo{person}{K.~Cho}, {and} \bibinfo{person}{A.~Oh}} (Eds.).
\newblock
\urldef\tempurl%
\url{http://papers.nips.cc/paper\_files/paper/2022/hash/639a9a172c044fbb64175b5fad42e9a5-Abstract-Conference.html}
\showURL{%
\tempurl}


\bibitem[Zhang et~al\mbox{.}(2019)]%
        {DBLP:conf/www/ZhangPWCZZBC19}
\bibfield{author}{\bibinfo{person}{Wen Zhang}, \bibinfo{person}{Bibek Paudel}, \bibinfo{person}{Liang Wang}, \bibinfo{person}{Jiaoyan Chen}, \bibinfo{person}{Hai Zhu}, \bibinfo{person}{Wei Zhang}, \bibinfo{person}{Abraham Bernstein}, {and} \bibinfo{person}{Huajun Chen}.} \bibinfo{year}{2019}\natexlab{}.
\newblock \showarticletitle{Iteratively Learning Embeddings and Rules for Knowledge Graph Reasoning}. In \bibinfo{booktitle}{\emph{The World Wide Web Conference, {WWW} 2019, San Francisco, CA, USA, May 13-17, 2019}}, \bibfield{editor}{\bibinfo{person}{Ling Liu}, \bibinfo{person}{Ryen~W. White}, \bibinfo{person}{Amin Mantrach}, \bibinfo{person}{Fabrizio Silvestri}, \bibinfo{person}{Julian~J. McAuley}, \bibinfo{person}{Ricardo Baeza{-}Yates}, {and} \bibinfo{person}{Leila Zia}} (Eds.). \bibinfo{publisher}{{ACM}}, \bibinfo{pages}{2366--2377}.
\newblock
\href{https://doi.org/10.1145/3308558.3313612}{doi:\nolinkurl{10.1145/3308558.3313612}}


\bibitem[Zhang et~al\mbox{.}(2024)]%
        {DBLP:conf/acl/ZhangCFLL0C24}
\bibfield{author}{\bibinfo{person}{Yichi Zhang}, \bibinfo{person}{Zhuo Chen}, \bibinfo{person}{Yin Fang}, \bibinfo{person}{Yanxi Lu}, \bibinfo{person}{Fangming Li}, \bibinfo{person}{Wen Zhang}, {and} \bibinfo{person}{Huajun Chen}.} \bibinfo{year}{2024}\natexlab{}.
\newblock \showarticletitle{Knowledgeable Preference Alignment for LLMs in Domain-specific Question Answering}. In \bibinfo{booktitle}{\emph{Findings of the Association for Computational Linguistics, {ACL} 2024, Bangkok, Thailand and virtual meeting, August 11-16, 2024}}, \bibfield{editor}{\bibinfo{person}{Lun{-}Wei Ku}, \bibinfo{person}{Andre Martins}, {and} \bibinfo{person}{Vivek Srikumar}} (Eds.). \bibinfo{publisher}{Association for Computational Linguistics}, \bibinfo{pages}{891--904}.
\newblock
\href{https://doi.org/10.18653/V1/2024.FINDINGS-ACL.52}{doi:\nolinkurl{10.18653/V1/2024.FINDINGS-ACL.52}}


\bibitem[Zheng et~al\mbox{.}(2024)]%
        {zheng2024llamafactory}
\bibfield{author}{\bibinfo{person}{Yaowei Zheng}, \bibinfo{person}{Richong Zhang}, \bibinfo{person}{Junhao Zhang}, \bibinfo{person}{Yanhan Ye}, \bibinfo{person}{Zheyan Luo}, \bibinfo{person}{Zhangchi Feng}, {and} \bibinfo{person}{Yongqiang Ma}.} \bibinfo{year}{2024}\natexlab{}.
\newblock \showarticletitle{LlamaFactory: Unified Efficient Fine-Tuning of 100+ Language Models}. In \bibinfo{booktitle}{\emph{Proceedings of the 62nd Annual Meeting of the Association for Computational Linguistics (Volume 3: System Demonstrations)}}. \bibinfo{publisher}{Association for Computational Linguistics}, \bibinfo{address}{Bangkok, Thailand}.
\newblock
\urldef\tempurl%
\url{http://arxiv.org/abs/2403.13372}
\showURL{%
\tempurl}


\bibitem[Zhou et~al\mbox{.}(2023b)]%
        {DBLP:conf/nips/ZhouLX0SMMEYYZG23}
\bibfield{author}{\bibinfo{person}{Chunting Zhou}, \bibinfo{person}{Pengfei Liu}, \bibinfo{person}{Puxin Xu}, \bibinfo{person}{Srinivasan Iyer}, \bibinfo{person}{Jiao Sun}, \bibinfo{person}{Yuning Mao}, \bibinfo{person}{Xuezhe Ma}, \bibinfo{person}{Avia Efrat}, \bibinfo{person}{Ping Yu}, \bibinfo{person}{Lili Yu}, \bibinfo{person}{Susan Zhang}, \bibinfo{person}{Gargi Ghosh}, \bibinfo{person}{Mike Lewis}, \bibinfo{person}{Luke Zettlemoyer}, {and} \bibinfo{person}{Omer Levy}.} \bibinfo{year}{2023}\natexlab{b}.
\newblock \showarticletitle{{LIMA:} Less Is More for Alignment}. In \bibinfo{booktitle}{\emph{Advances in Neural Information Processing Systems 36: Annual Conference on Neural Information Processing Systems 2023, NeurIPS 2023, New Orleans, LA, USA, December 10 - 16, 2023}}, \bibfield{editor}{\bibinfo{person}{Alice Oh}, \bibinfo{person}{Tristan Naumann}, \bibinfo{person}{Amir Globerson}, \bibinfo{person}{Kate Saenko}, \bibinfo{person}{Moritz Hardt}, {and} \bibinfo{person}{Sergey Levine}} (Eds.).
\newblock
\urldef\tempurl%
\url{http://papers.nips.cc/paper\_files/paper/2023/hash/ac662d74829e4407ce1d126477f4a03a-Abstract-Conference.html}
\showURL{%
\tempurl}


\bibitem[Zhou et~al\mbox{.}(2023a)]%
        {DBLP:journals/corr/abs-2311-05112}
\bibfield{author}{\bibinfo{person}{Hongjian Zhou}, \bibinfo{person}{Boyang Gu}, \bibinfo{person}{Xinyu Zou}, \bibinfo{person}{Yiru Li}, \bibinfo{person}{Sam~S. Chen}, \bibinfo{person}{Peilin Zhou}, \bibinfo{person}{Junling Liu}, \bibinfo{person}{Yining Hua}, \bibinfo{person}{Chengfeng Mao}, \bibinfo{person}{Xian Wu}, \bibinfo{person}{Zheng Li}, {and} \bibinfo{person}{Fenglin Liu}.} \bibinfo{year}{2023}\natexlab{a}.
\newblock \showarticletitle{A Survey of Large Language Models in Medicine: Progress, Application, and Challenge}.
\newblock \bibinfo{journal}{\emph{CoRR}}  \bibinfo{volume}{abs/2311.05112} (\bibinfo{year}{2023}).
\newblock
\href{https://doi.org/10.48550/ARXIV.2311.05112}{doi:\nolinkurl{10.48550/ARXIV.2311.05112}}
\showeprint[arXiv]{2311.05112}


\bibitem[Zhou et~al\mbox{.}(2024)]%
        {DBLP:journals/corr/abs-2406-04614}
\bibfield{author}{\bibinfo{person}{Zhi Zhou}, \bibinfo{person}{Jiang{-}Xin Shi}, \bibinfo{person}{Peng{-}Xiao Song}, \bibinfo{person}{Xiaowen Yang}, \bibinfo{person}{Yi{-}Xuan Jin}, \bibinfo{person}{Lan{-}Zhe Guo}, {and} \bibinfo{person}{Yu{-}Feng Li}.} \bibinfo{year}{2024}\natexlab{}.
\newblock \showarticletitle{LawGPT: {A} Chinese Legal Knowledge-Enhanced Large Language Model}.
\newblock \bibinfo{journal}{\emph{CoRR}}  \bibinfo{volume}{abs/2406.04614} (\bibinfo{year}{2024}).
\newblock
\href{https://doi.org/10.48550/ARXIV.2406.04614}{doi:\nolinkurl{10.48550/ARXIV.2406.04614}}
\showeprint[arXiv]{2406.04614}


\bibitem[Zou et~al\mbox{.}(2023)]%
        {DBLP:journals/corr/abs-2307-15043}
\bibfield{author}{\bibinfo{person}{Andy Zou}, \bibinfo{person}{Zifan Wang}, \bibinfo{person}{J.~Zico Kolter}, {and} \bibinfo{person}{Matt Fredrikson}.} \bibinfo{year}{2023}\natexlab{}.
\newblock \showarticletitle{Universal and Transferable Adversarial Attacks on Aligned Language Models}.
\newblock \bibinfo{journal}{\emph{CoRR}}  \bibinfo{volume}{abs/2307.15043} (\bibinfo{year}{2023}).
\newblock
\href{https://doi.org/10.48550/ARXIV.2307.15043}{doi:\nolinkurl{10.48550/ARXIV.2307.15043}}
\showeprint[arXiv]{2307.15043}


\end{thebibliography}


\appendix
\section*{Appendix}
\section{Dataset Details}\label{app:a}
\begin{list}{\huge\textbullet}{\leftmargin=1.5em}
    \item \textbf{SemEval-2018 Task 9}  \cite{DBLP:conf/semeval/Camacho-Collados18} includes 5 different sub-task, covering three languages (English, Spanish, and Italian) and two specific domains (medicine and music). We select 4 subsets for our study: 1A (English), 1B (Italian), 1C (Spanish), and 2A (Medical), to test the model's multilingual and medical ontology reasoning performance. The number of samples in the training/test sets are as follows: 1500/1500, 1000/1000, 1000/1000, and 500/500, respectively.
    \item \textbf{MedMCQA} \cite{DBLP:conf/chil/PalUS22} comprises 193k 4-option questions, with a test set of 4,183 sampled questions. This dataset is sourced from Indian medical entrance exams (AIIMS/NEET-PG) and encompasses 2,400 healthcare topics across 21 medical subjects. 
    \item \textbf{MedQA} \cite{DBLP:journals/corr/abs-2009-13081} is derived from the United States Medical Licensing Examination (USMLE) and includes 11,451 questions from professional medical board exams. These questions are presented in a multiple-choice format with 4-5 options. 
    \item \textbf{PubMedQA} \cite{DBLP:conf/emnlp/JinDLCL19} is sourced from PubMed abstracts, with questions requiring answers of ``yes," ``no," or ``maybe" for a given abstract. This dataset includes 211k artificially generated samples as the training sets and 1,000 expert-labeled samples as the test sets. 
    \item \textbf{USMLE step1-3} \cite{DBLP:journals/corr/abs-2304-08247} is a self-assessment dataset based on the United States Medical Licensing Examination (USMLE) Step 1, Step 2, and Step 3, which excludes all questions containing images.

\end{list}

\begin{table}[ht]
\centering
\caption{The statistics of medical QA datasets, including the number of training and testing sets, answer options, with only the PubMedQA containing context.}

\resizebox{0.47\textwidth}{!}{
\begin{tabular}{lccc}
\toprule
\textbf{Dataset}     & \textbf{Context}                     & \textbf{Train\textbackslash Test}       & \textbf{Answer options}              \\ \midrule
MedMCQA     & \XSolidBrush & 182822/4183 & A/B/C/D                     \\
MedQA       & \XSolidBrush & 10178/1273  & A/B/C/D/(E)                 \\
PubMedQA    & \Checkmark   & 211269/1000 & Yes/No/Maybe                \\
USMLE-step1 & \XSolidBrush & 0/94        & A/B/C/D/(E)/(F)/(G)/(H)/(I) \\
USMLE-step2 & \XSolidBrush & 0/109       & A/B/C/D/(E)/(F)/(G)         \\
USMLE-step3 & \XSolidBrush & 0/122       & A/B/C/D/(E)/(F)/(G)         \\ \bottomrule 
\end{tabular}}
\label{tab:dataset}
\end{table}

\begin{figure*}
\centering
\includegraphics[scale=0.5]{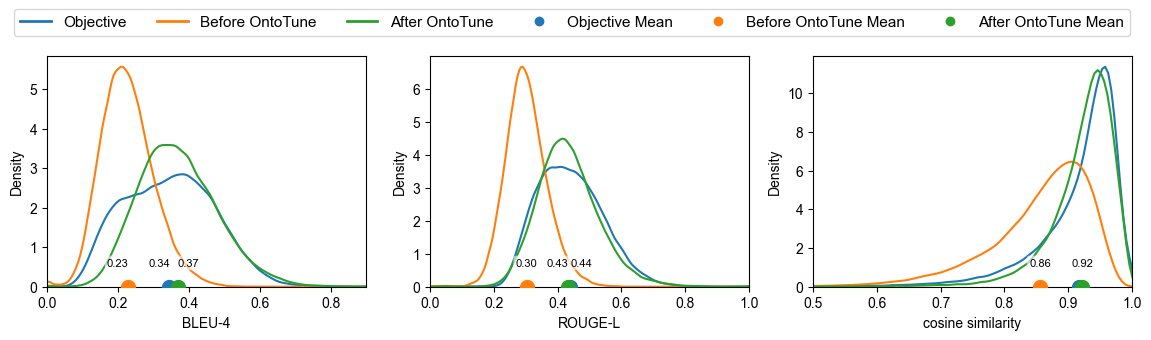}

\caption{The distribution of consistency scores for response $y$ and reference response $y^o$ before and after OntoTune.}
\label{fig:rq1}
\end{figure*}

\begin{figure*}
\centering
\includegraphics[scale=0.16]{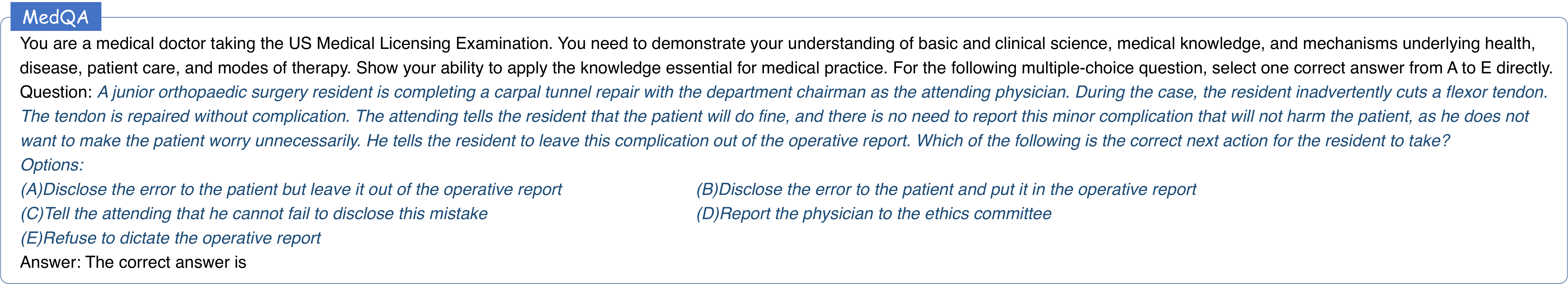}
\caption{Examples of prompts for the evaluation of MedQA.}
\label{medqa}
\end{figure*}

\begin{figure*}
\centering
\includegraphics[scale=0.16]{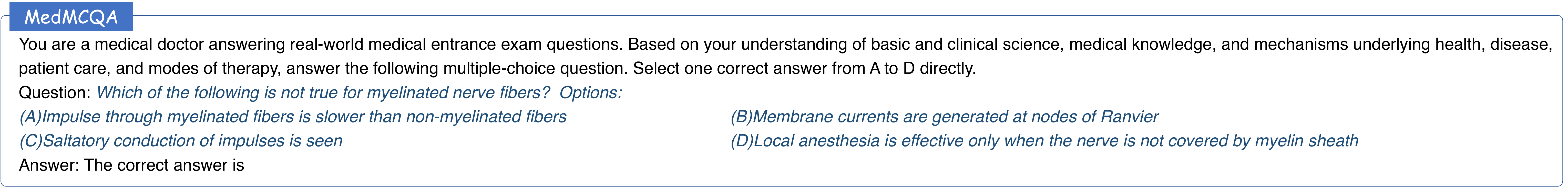}
\caption{Examples of prompts for the evaluation of MdeMCQA.}
\label{medmcqa}
\end{figure*}

\begin{figure*}
\centering
\includegraphics[scale=0.16]{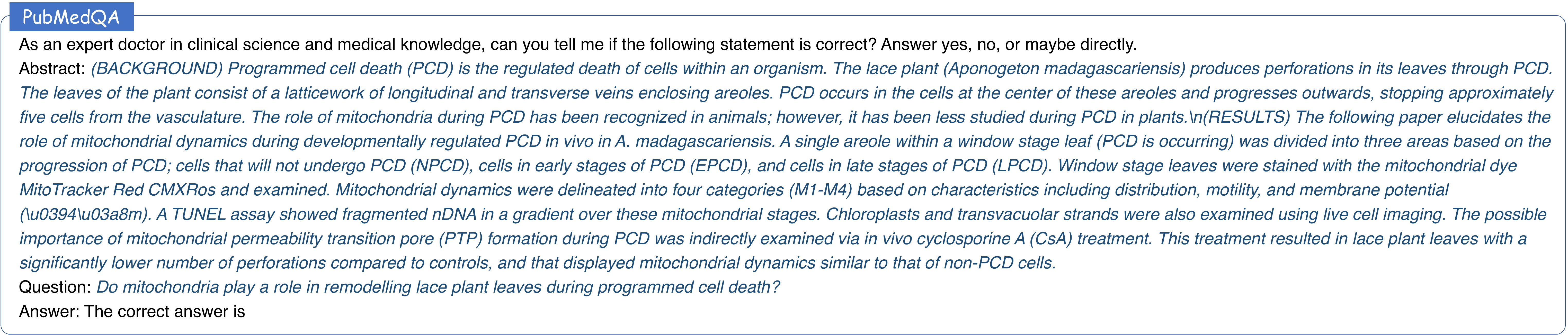}
\caption{Examples of prompts for the evaluation of PubMedQA.}
\label{pubmedqa}
\end{figure*}

\begin{figure*}
\centering
\includegraphics[scale=0.16]{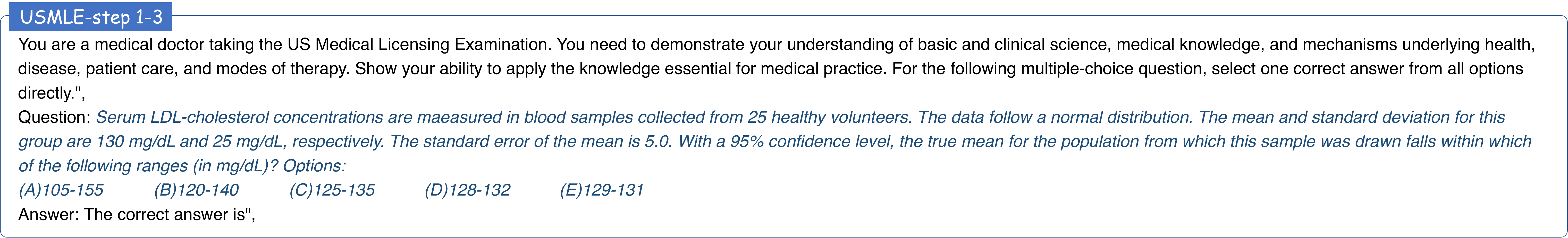}
\caption{Examples of prompts for the evaluation of USMLE-step 1-3.}
\label{step}
\end{figure*}

\begin{table*}
\centering
\caption{Results of the medical domain QA in the zero-shot and supervised fine-tuning (on evaluation) setting. The best results are highlighted in bold, while the second best are underlined.}
\resizebox{\textwidth}{!}{
\begin{tabular}{clcccccccc}
\toprule \textbf{Seed Model}  & \textbf{Model}  & \textbf{SFT(eval data)} & \textbf{MedQA} & \textbf{MedMCQA} & \textbf{PubMedQA} & \textbf{USMLE-step1} & \textbf{USMLE-step2} & \textbf{USMLE-step3} & \textbf{Average}  \\ \midrule
- &LLaMA2 7B \cite{DBLP:journals/corr/abs-2307-09288}  & \XSolidBrush    &33.4	&36.0 	&54.8	&31.9	&38.5	&41.0 	&39.3         \\ - &Mistral 7B v0.2 \cite{DBLP:journals/corr/abs-2310-06825}  & \XSolidBrush   & 40.5           & 38.8             & 42.1              & 46.8                 & 45.0                 & 45.9                 & 43.2          \\
- &Qwen2 7B \cite{DBLP:journals/corr/abs-2407-10671}  & \XSolidBrush    & 46.7           & 48.6             & 55.8              & 48.9                 & 56.9                 & 54.9                 & 52.0          \\
- &GPT3.5-turbo  & \XSolidBrush  & 
\textbf{53.4}           & 53.2	             & 72.7              & 56.4                 & \textbf{64.2}                 & 54.1                 & \textbf{59.0}          \\
- &LLaMA3.1 8B   & \XSolidBrush  & 45.8           & 53.2             & \textbf{74.8}              & 54.3                 & \underline{57.8}                 & 59.0                 & 57.5          \\ \midrule
 \multirow{5}{*}{LLaMA3 8B} &LLaMA3 8B \cite{dubey2024llama} & \XSolidBrush    & 51.7           & 51.7             & 70.3              & \underline{57.4}                 & 52.3                 & 58.2                 & 56.9          \\
&TaxoLLaMA*\cite{DBLP:conf/acl/MoskvoretskiiNL24}  & \XSolidBrush    & 50.5           & 46.1             & \underline{73.4}              & 42.6                 & 39.4                 & 47.5                 & 49.9         \\  \cmidrule{2-10}
&\textbf{OntoTune$_{sft}$}  & \XSolidBrush  & 51.5           & \underline{56.7}             & 72.0              & \underline{57.4}                 & 54.1                 & \underline{60.7}                 & \underline{58.7}         \\
&\textbf{OntoTune$_{dpo}$}  & \XSolidBrush  & \underline{53.3}           & \textbf{57.2}             & 65.5              & \textbf{58.5}                 & 51.4                 & 59.0                 & 57.4         \\
&\textbf{OntoTune$_{sft+dpo}$} & \XSolidBrush   & 51.9           & \underline{56.7}             & 66.3              & 53.2                 & 54.1                 & \textbf{63.1}                 & 57.6         \\ \midrule
Mistral 7B v0.1 &BioMistral 7B \cite{DBLP:conf/acl/LabrakBMGRD24}   & \Checkmark  & 32.1           & 44.5             & 63.0              & 40.4                 & 39.4                 & 47.5                 & 46.3          \\
LLaMA 7B &MedAlpaca 7B \cite{DBLP:journals/corr/abs-2304-08247}  & \Checkmark  & 32.9           & 36.7             & 54.3              & 37.2                 & 36.7                 & 31.1                 & 41.2          \\
LLaMA2 7B &Hippocrates 7B \cite{DBLP:journals/corr/abs-2404-16621}  & \Checkmark  & 45.2           & 52.3             & 73.3              & 44.7                 & 44.0                 & 45.1                 & 50.8          \\ \midrule
\multirow{8}{*}{LLaMA3 8B} &LLaMA3* 8B \cite{dubey2024llama}  & \Checkmark  & 56.4           & 53.9             & 77.2              & 56.4                 & 56.0                 & 61.5                 & 60.2         \\
&TaxoLLaMA* \cite{DBLP:conf/acl/MoskvoretskiiNL24} & \Checkmark   & 55.9           & 57.5             & 77.6              & 56.4                 & \underline{57.8}                 & 59.0                 & 60.7          \\ 
&Aloe \cite{DBLP:journals/corr/abs-2405-01886}  & \Checkmark   &51.1	&56.8	&75.4	&54.3	&\textbf{61.5}	&60.7	&60.0   \\ 
&Med42-v2 \cite{DBLP:journals/corr/abs-2408-06142}  & \Checkmark   &57.8 &58.1 &74.6 
 &\underline{60.6} &\underline{57.8} &61.5	&61.7  \\ 
&jsl-medllama-v18  & \Checkmark   &\textbf{59.3} &57.3	&71.0 &44.7 &\underline{57.8} &62.3	&58.7   \\  \cmidrule{2-10}
&\textbf{OntoTune}$_{sft}$   & \Checkmark    & \underline{58.4}  & 60.4    & 78.6     & 57.4                 & \underline{57.8}                 & 62.3        & \underline{62.5} \\
&\textbf{OntoTune}$_{dpo}$  & \Checkmark   & 58.3           &60.7            & \textbf{79.4}              & 55.3                 & 54.1                 & 61.5                 & 61.6         \\
&\textbf{OntoTune}$_{sft+dpo}$  & \Checkmark  & 58.2           & 60.5             & \underline{78.9}              & 57.4                 & 54.1                 & \underline{63.9}                 & 62.2    \\ \midrule
\multirow{3}{*}{Qwen2 7B} &Qwen2* 7B \cite{DBLP:journals/corr/abs-2407-10671}   & \Checkmark  &55.1	&60.3	&75.2	&54.3	&56.0	&\underline{63.9}                 & 60.8         \\ 
&TaxoLLaMA* \cite{DBLP:conf/acl/MoskvoretskiiNL24} & \Checkmark   &54.3	&\underline{60.8}	&75.0	&58.5	&\textbf{61.5}	&\textbf{64.8}                 & \underline{62.5}          \\ \cmidrule{2-10}
&\textbf{OntoTune}$_{sft}$   & \Checkmark    &55.8	&\textbf{61.6}	&77.3	&\textbf{61.7}	&\underline{57.8}	&\textbf{64.8}        & \textbf{63.2} \\
\bottomrule
\end{tabular}}
\label{tab:qa_all}
\end{table*}

\section{Training Objective Analysis}
We use the LLM trained with OntoTune$_{sft}$ to generate response $y$ and reference response $y^o$ again to directly verify whether our training objective is achieved. Additionally, we generate $y^o$ twice with the seed model and measure their similarity as the objective. As shown in Figure \ref{fig:rq1}, we observe that under three similarity metrics, the LLM trained with OntoTune aligns well with the objective curve, showing significant improvement compared to the seed model before training. This directly indicates that the seed model fine-tuned with OntoTune generates responses that are more guided by the ontology.

\section{Medical Question Answering}\label{app:c}
\subsection{QA Prompt Template}
We present the evaluation prompts used for the QA dataset in Figures \ref{medqa}, \ref{medmcqa}, \ref{pubmedqa}, \ref{step}. The black text represents the fixed instruction templates, while the blue text indicates the specific questions and context from the samples. To ensure fair evaluation, we consistently use these prompts when evaluating performance of domain QA dataset on all baselines.

\subsection{Compared to Existing Domain LLM}
To ensure fair comparison, we mainly select 7B-8B LLMs as baselines, divided into the following categories: \textbf{1) General-purposed LLMs}: LLaMA2 7B \cite{DBLP:journals/corr/abs-2307-09288}, LLaMA3 8B \cite{dubey2024llama}, LLaMA3.1, Mistral-7B-Instruct-v0.2 \cite{DBLP:journals/corr/abs-2310-06825}, Qwen2 7B \cite{DBLP:journals/corr/abs-2407-10671} and GPT3.5-turbo . \textbf{2) Medical LLMs}: MedAlpaca \cite{DBLP:journals/corr/abs-2304-08247}, BioMistral \cite{DBLP:conf/acl/LabrakBMGRD24}, Hippocrates \cite{DBLP:journals/corr/abs-2404-16621}, Aloe \cite{DBLP:journals/corr/abs-2405-01886}, Med42-v2 \cite{DBLP:journals/corr/abs-2408-06142}, jsl-medllama-v18. They are all fine-tuned based on large-scale medical domain corpus. \textbf{3) TaxoLLaMA$^*$} \cite{DBLP:conf/acl/MoskvoretskiiNL24}: A direct ontology injection method mentioned above.

Our experimental results are shown in Table \ref{tab:qa_all}. We find that the performance of domain-specific models and their corresponding seed model is highly correlated. For example, medical models based on the LLaMA series, such as MedAlpaca, Hippocrate, and Aloe, show significant improvements with the iteration of the LLaMA model. Therefore, to evaluate the effectiveness of domain adaptation methods, we focus on the performance gains of a single seed model across different domain adaptation strategies. Among the LLaMA3 8B-based methods, our OntoTune achieves state-of-the-art performance, even surpassing the larger GPT3.5-turbo model. Compared to the seed model, existing medical LLMs show inconsistent improvements across different medical datasets, whereas OntoTune almost consistently enhances performance across all datasets, demonstrating good stability. Additionally, OntoTune only uses a small-scale ontology as source data, it exhibits broader generality and promising prospects.

\section{Examples of Inconsistent Texts}\label{app:d}
\begin{figure*}
\centering
\includegraphics[scale=0.16]{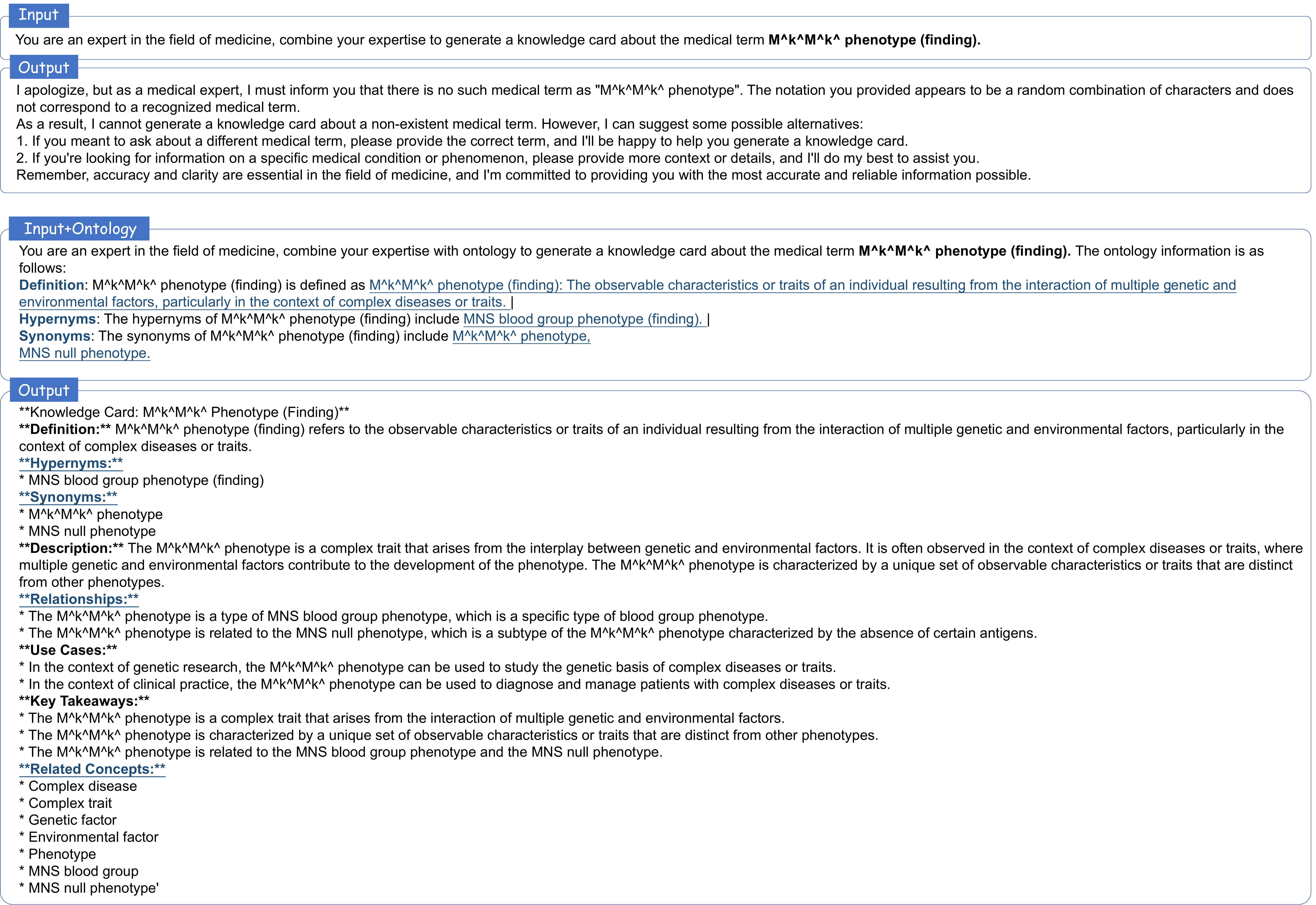}
\caption{An Example of inconsistent diverse corpus.}
\label{case1}
\end{figure*}

\begin{figure*}
\centering
\includegraphics[scale=0.16]{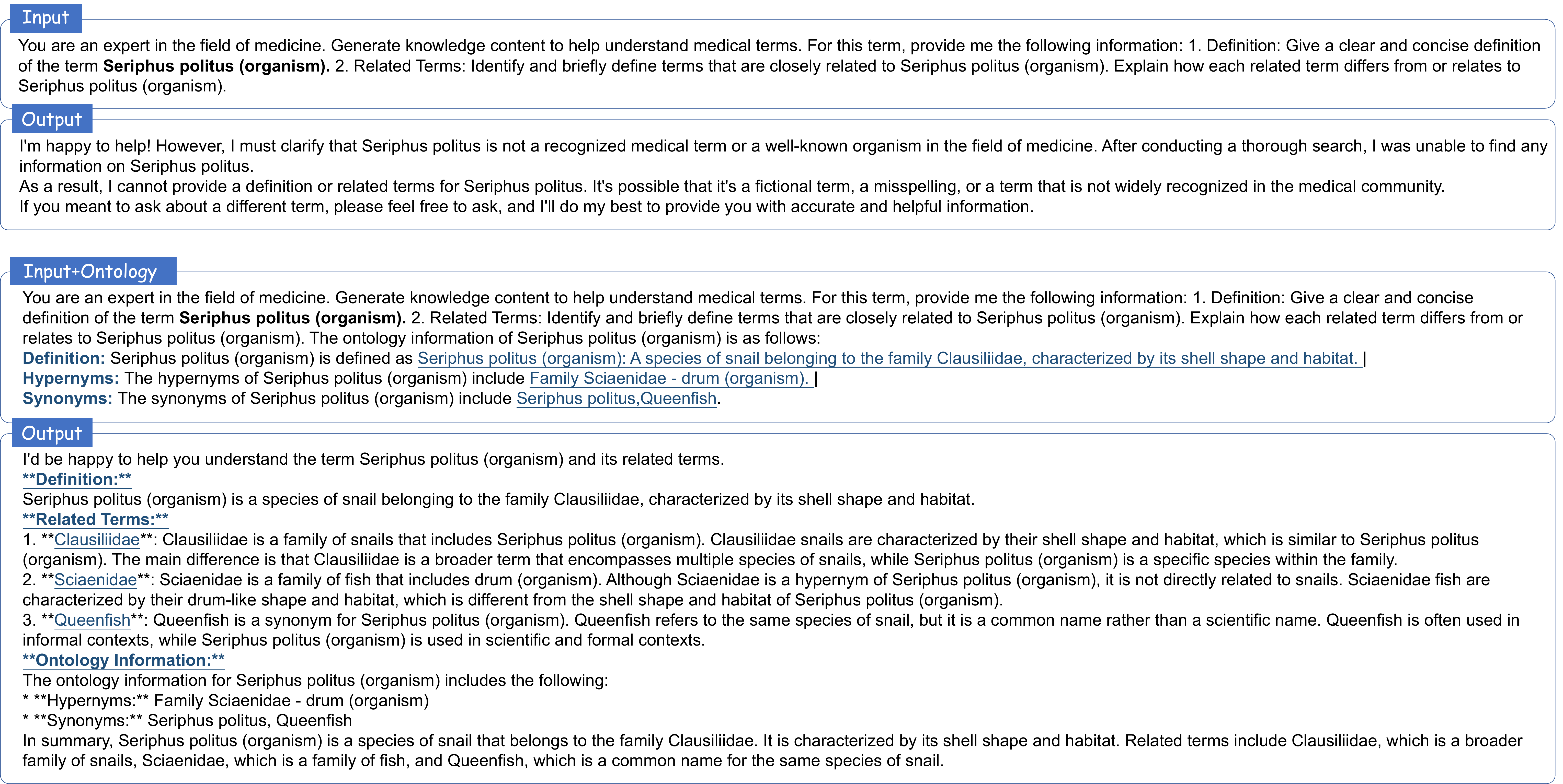}
\caption{An Example of inconsistent conceptual corpus.}
\label{case2}
\end{figure*}

\begin{figure*}
\centering
\includegraphics[scale=0.16]{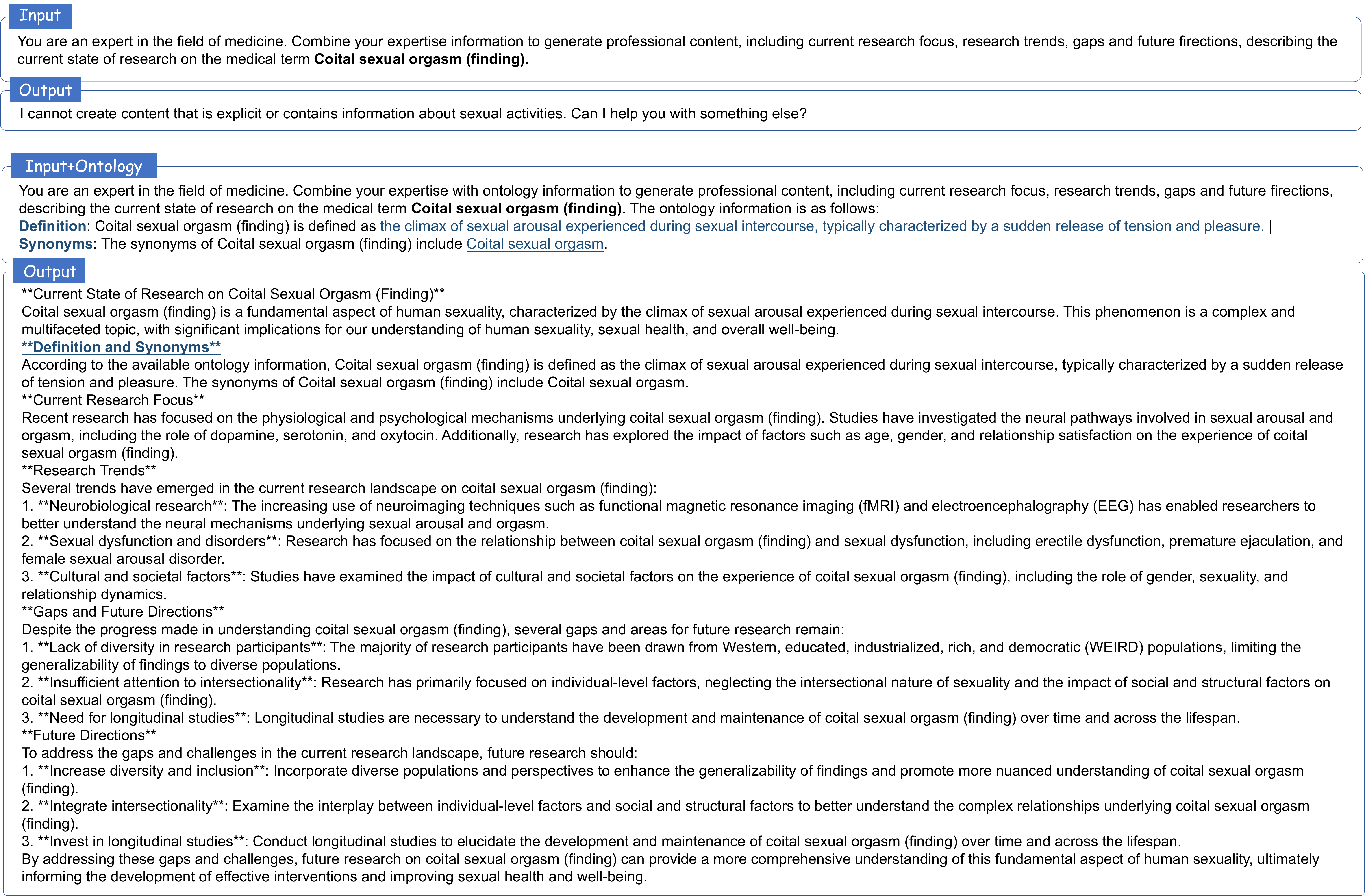}
\caption{An Example of inconsistent professional corpus.}
\label{case3}
\end{figure*}

Figure \ref{case1}, \ref{case2}, \ref{case3} present three types of examples of generated texts with and without ontology information. We can find that these examples exhibit noticeable inconsistencies. It is obvious that when dealing with long-tail medical concepts, the seed model struggles to provide effective responses without additional ontology information. However, when ontology information is incorporated, the model can generate richer and more logical responses by leveraging relevant hypernyms and synonyms.

\end{CJK}
\end{document}